\documentclass[10pt,twocolumn,letterpaper]{article}

\usepackage[pagenumbers]{cvpr}
\usepackage{bm}
\usepackage{multirow}
\usepackage{makecell}
\usepackage{colortbl}
\usepackage{tabularx}
\usepackage{adjustbox}
\usepackage{placeins}

\definecolor{best}{RGB}{234,239,239}

\definecolor{cvprblue}{rgb}{0.21,0.49,0.74}
\usepackage[breaklinks,colorlinks,allcolors=cvprblue]{hyperref}

\title{ERF-GS: Reconstructing Fast Motion from Disjoint Event-RGB Viewpoints}

\author{
Xiaoyang Bai$^{1,*}$ \quad Zhenyang Li$^{1,*}$ \quad Weiwei Xu$^{2}$\\
Edmund Y. Lam$^{1,\dagger}$ \quad Yifan Peng$^{1,\dagger}$\\
$^{1}$Department of Electrical and Electronic Engineering,\\
The University of Hong Kong, Hong Kong SAR, China\\
$^{2}$State Key Lab of CAD\&CG, Zhejiang University, Hangzhou, China\\
{\tt\small xybai@hku.hk; lizy23@connect.hku.hk; xww@cad.zju.edu.cn}\\
{\tt\small elam@eee.hku.hk; evanpeng@hku.hk}\\
$^{*}$Equal contribution. $^{\dagger}$Corresponding authors.
}

\begin{document}
\maketitle

\begin{abstract}
Deep learning-driven representations such as neural radiance fields (NeRFs)
and 3D Gaussian splatting (3DGS) have revolutionized the field of dynamic 3D
scene reconstruction with improved visual precision and scalability. However,
the reconstruction of fast-moving objects remains a challenge; existing
methods based on conventional frame-based videos often struggle in scenarios
such as sports events and animal videography. We propose an event-RGB fusion
Gaussian splatting (ERF-GS) framework that integrates event information into
both optimization and densification stages of the Gaussian splatting pipeline,
taking advantage of novel event sensors with high frame-rate. Unlike many other
event-assisted scene reconstruction methods, ERF-GS was developed using
realistic simulation settings and realizes event-based learning detached from
RGB inputs. This design enables its application beyond straightforward
synthetic data into the realm of natural video with complex layout, low frame
rates and severe motion blur. Our experiments show that ERF-GS outperforms both
the 4DGS baseline and the concurrent E-D3DGS on different variants of the
Neu3D and Nvidia datasets which include blurry RGB frames and disjoint
RGB-event viewpoints. Our code is available at
\url{https://github.com/andrewbxy/ERF-GS}.
\end{abstract}


\section{Introduction}
\label{sec:intro}

\begin{figure*}[t!]
  \includegraphics[width=\textwidth]{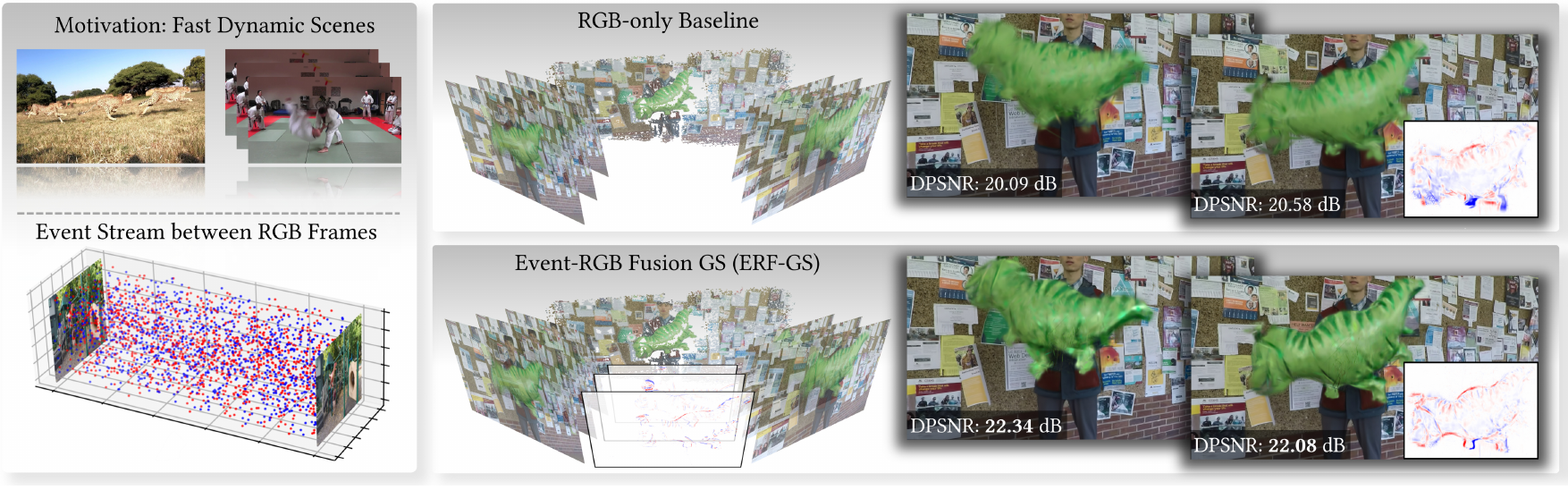}
   \caption{Top left: real-world applications such as sports events and animal videography challenge existing dynamic scene reconstruction algorithms due to fast object motion and blurred RGB frames. Bottom left: the high temporal resolution of event data makes it an ideal choice for multimodal fusion-based solutions. Right: compared to an RGB-only baseline, our ERF-GS framework achieves higher-quality reconstruction of fast real-world dynamics using blurred RGB frames and separate RGB-event viewpoints.}
   \label{fig:teaser}
\end{figure*}

Recent years have seen a growth in deep learning-based 3D scene reconstruction methods, pioneered by two outstanding representations: neural radiance fields (NeRFs)~\cite{mildenhall2021nerf} and 3D Gaussian splatting (3DGS)~\cite{kerbl20233d}. Powerful as these methods are, they are as yet unable to model dynamic 3D scenes and thus cannot meet the needs of the growing AR/VR and robotics industries~\cite{wang2024nerf}, or support downstream tasks such as motion synthesis and generation~\cite{hou2024causal}. 
Motivated by such demands, researchers have attempted to extend both NeRFs~\cite{pumarola2021d, li2023dynibar} and 3DGS~\cite{wu20244d, li2024spacetime} to  dynamic settings. However, challenges such as motion blur and large deformation between frames are still largely unresolved, precluding their application to scenarios such as sports events or animal videography (see Fig.~\ref{fig:teaser}(top left)). Since artifacts in these scenes are predominantly motion-related and arise due to the inherently long exposure times of frame-based cameras, resolving them within the existing  setting which solely uses RGB (intensity) images as  input can be an arduous endeavor.

The invention of event sensors~\cite{gallego2020event} has ushered in new opportunities for multimodal fusion solutions. Unlike conventional sensors that capture dense, absolute light intensities as pixel values, event sensors record sparse and asynchronous signal spikes (`events') that represent relative changes in light intensity at each pixel location, as indicated in Fig.~\ref{fig:teaser}(bottom left). They exhibit low latency, low power requirements, and high sensitivity to motion, but suffer from  high noise  and a lack of texture and color information, making them an ideal complement to RGB images that are rich in dense visual details but struggle with fast object movements. To date, event sensors have been employed to help in various computer vision tasks such as classification~\cite{deng2022voxel}, tracking~\cite{zhang2021object}, HDR imaging~\cite{messikommer2022multi}, motion prediction~\cite{monforte2020exploiting}, and intelligent camera control~\cite{louAllinFocusImagingEvent2023,lin2024embodied}, demonstrating their potential in both low-level and high-level visual computing applications.

Inspired by this progress, researchers have recently started utilizing event information for both static~\cite{deguchi2024e2gs, xiong2024event3dgs} and dynamic~\cite{ma2023deformable, xu2025event} scene reconstruction. However, the gap between such work and real-world applications persists due to the following issues that are largely overlooked. 
\begin{itemize}
\item \emph{Event-RGB commonality:} almost all existing research requires the event and RGB inputs to share the same camera viewpoint or the same number of channels, which are hard to achieve with current hardware designs. These commonalities between the two modalities also restricts their robustness to motion blur. 
\item \emph{Idealized experimental setups:} most works are trained and evaluated on synthetic scenes or natural videos without motion-induced artifacts; their performance under more challenging and realistic conditions has not been tested. 
\item \emph{Lack of event-based density control:} the majority of existing methods focus on using event information to construct loss terms, deblur RGB frames, or estimate camera poses, while another important aspect of Gaussian splatting, density control, is left unconsidered.
\end{itemize}

To address these issues, we propose an event-RGB fusion Gaussian splatting  framework, ERF-GS, that features two unique components, an \emph{event-assisted regularized loss (EARL)} and an \emph{event-guided densification strategy (EDS)}, that update the dynamic Gaussian representation with event information at different stages of the learning pipeline. We prioritize the addressing of real-world constraints in our algorithm design. Thus, ERF-GS realizes event-based learning that does not require ground truth RGB frames or event-RGB alignment. This not only mitigates the impact of motion-induced artifacts in input video, thus enabling our proposed framework to function in more challenging conditions, but it also allows ERF-GS to be incorporated into any existing dynamic Gaussian splatting method without modifying its Gaussian representation or core deformation algorithm, allowing it wide application to downstream use cases.

Instead of resorting to purely synthetic data, we employ the established v2e algorithm~\cite{hu2021v2e} to simulate event streams from existing real-world datasets such as Nvidia~\cite{yoon2020novel} and Neu3D~\cite{li2022neural}. In order to test our model with fast scene dynamics, we temporally subsample each RGB video and degrade the resulting RGB frames with artificial motion blur. Moreover, we use disjoint sets of viewpoints for RGB and event inputs so that the curated datasets closely mimic the real-world capture of rapidly moving objects with unaligned RGB and event cameras. Experimental results show that our ERF-GS framework  improves novel view synthesis (NVS) quality  by $>0.9$~dB  dynamic PSNR when applied to the state-of-the-art 4DGS~\cite{wu20244d} baseline (see Fig.~\ref{fig:teaser}{right}), extending the scope of event-based dynamic scene reconstruction to  more challenging and realistic cases. 
We summarize our contributions as follows:
\begin{itemize}
\item \emph{ERF-GS}, an event-RGB fusion pipeline that effectively leverages the high temporal resolution of event streams to enhance the reconstruction of fast-moving 3D scenes in realistic setups.
\item EARL and EDS, two innovative components which fuse event information into the GS pipeline without requiring ground truth RGB frame supervision or precise event-RGB alignment. This facilitates the application of ERF-GS to any existing dynamic GS baseline with minimal modification. 
\item Challenging datasets curated from popular benchmarks that closely mimic real-world scenarios with fast object motion and unaligned RGB and event cameras. Our model demonstrates visible improvements against the unimodal RGB baseline on these datasets.
\end{itemize}

\paragraph{Code and dataset release.}
The source code is available at \url{https://github.com/andrewbxy/ERF-GS}. Our processed Neu3D and Nvidia datasets are released at \url{https://huggingface.co/datasets/andrewbxy/ERFGS-Neu3D} and \url{https://huggingface.co/datasets/andrewbxy/ERFGS-Nvidia}, respectively.

Our work is one of the earliest attempts to improve dynamic scene reconstruction using event inputs, and it is inevitably limited in several ways. For instance, since no public dataset featuring real-world multiview event-RGB videos is available, we have had to rely on simulated events for all experiments, as have previous works such as DE-NeRF~\cite{ma2023deformable}, Dynamic EventNeRF~\cite{rudnev2024dynamic}, and event-boosted 3DGS~\cite{xu2025event}. Moreover, our experimental setting assumes a relatively large number of static cameras ($>10$), which is not easily attainable in the real world. 
Despite these shortcomings, we have designed our method and data for as  realistic setting as  possible, and  hope this work  serves to inspire future work that  mitigates these issues. 
 Sec.~\ref{sec:conclusion} provides a more detailed discussion, where we consider real-world data collection and generalizing ERF-GS to more realistic settings in future.

\begin{figure*}[t]
  \centering
   \includegraphics[width=\linewidth]{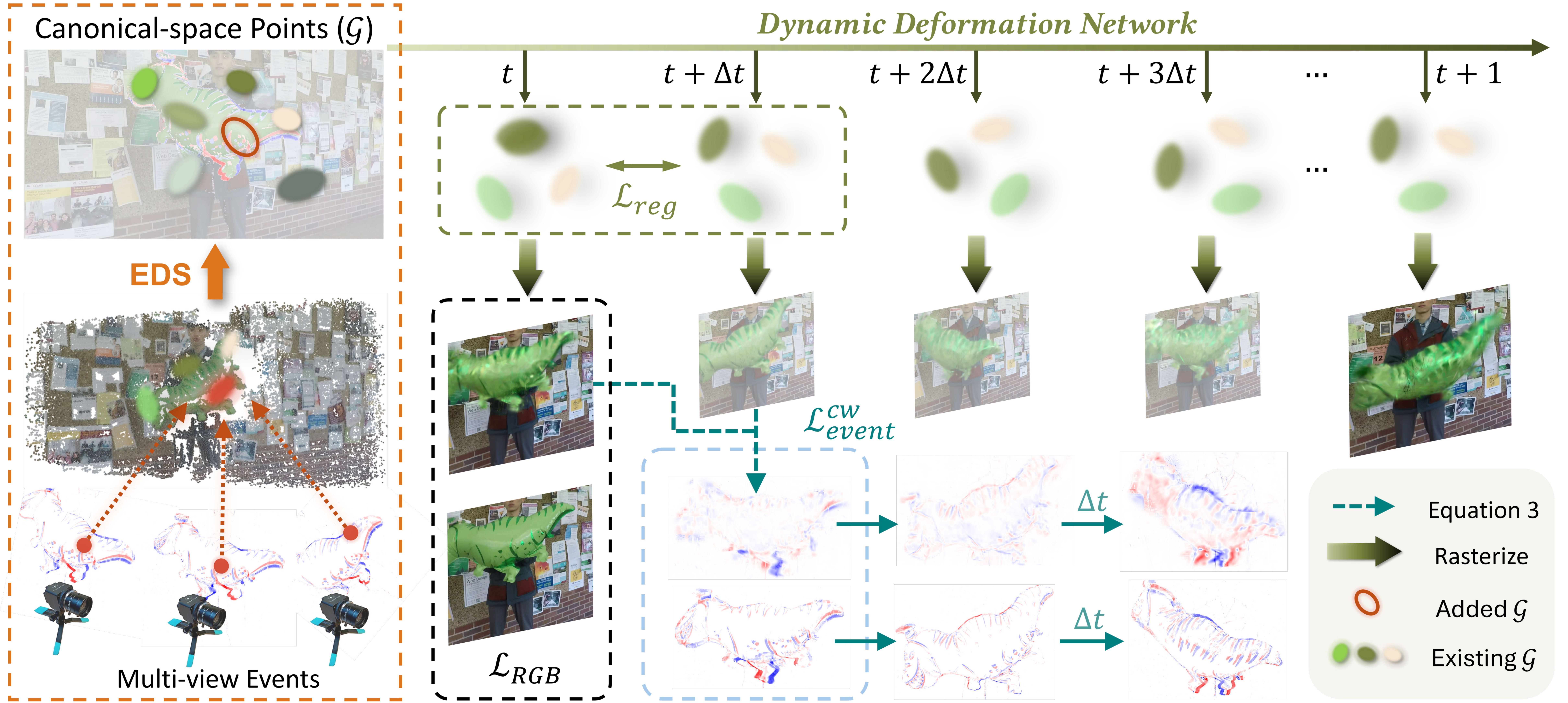}
   \caption{Overview of our ERF-GS framework. Event data is incorporated into the dynamic Gaussian splatting pipeline using two components: {EARL} supervises the deformed Gaussian representation between two RGB frames by calculating {$\mathcal{L}_\mathrm{event}^{\mathrm{cw}}$} based on rendered images, and constraining the learning procedure with trajectory and color regularization {$\mathcal{L}_\mathrm{reg}$}.  {EDS} locates movements in the 3D space through multiview event cues and directly adds Gaussians to those regions. The functionality of EARL and EDS does not depend on ground truth RGB frames, which are prone to motion-induced artifacts. 
   }
   \label{fig:pipeline}
\end{figure*}

\section{Related Work}

\subsection{Dynamic Scene Reconstruction}
Dynamic 3D reconstruction~\cite{ingale2021real} is a well-established topic in the field of visual computing. While past works based on surface fusion \cite{newcombe2015dynamicfusion} or non-rigid correspondence matching \cite{bozic2020deepdeform} have proven the feasibility of recovering 3D models from multiview videos, their use is nonetheless limited to simple, background-free scenes, thereby restricting their application to the real-world.

Recent progresses on NeRF-based \cite{mildenhall2021nerf} and 3DGS-based \cite{kerbl20233d} methods have largely removed this bottleneck~\cite{wu2024recent}. In the former category, scene dynamics is usually modeled through a time-conditional radiance field~\cite{du2021neural, xian2021space}. Building on this idea, D-NeRF~\cite{pumarola2021d} learns a deformation network alongside the canonical radiance field. DynamicNeRF~\cite{gao2021dynamic} regresses optical flow from spacetime coordinates. Nerfie~\cite{park2021nerfies} and HyperNeRF~\cite{park2021hypernerf} enhance the neural module with hyperspace representations. DynIBaR~\cite{li2023dynibar} models the trajectory field with a DCT basis. 
DMiT~\cite{yang2024dmit} leverages triplane representation to reduce computation time. 
However, training NeRF variants still demands substantial computational resources, and the resulting representations often lack object awareness and explainability.

3DGS and its variants~\cite{ren2024octree, yu2024mip} feature accelerated training speed and reduced memory usage compared to NeRF-based models, with recent works accelerating the pipeline to $>150$ fps for static scenes~\cite{peng2025gaussian}. For video reconstruction, dynamic 3DGS~\cite{luiten2023dynamic} equips Gaussians with per-frame parameters. Spacetime Gaussian~\cite{li2024spacetime} further generalizes that to continuous representations. Deformable 3DGS~\cite{yang2024deformable} trains a deformation field to model object motion; 4DGS~\cite{wu20244d} improves it with HexPlane representation~\cite{cao2023hexplane}. GaussianFlow~\cite{gao2024gaussianflow} incorporates optical flow for smoother trajectory learning. Finally, SC-GS~\cite{huang2024sc} introduces sparse control points to realize scene editing. With those methods, state-of-the-art reconstruction quality has been achieved on various datasets with complex scene layouts and object movements.

Despite their impressive results, all of these  methods suffer from output degradation when fast-moving objects in the scene cause motion blur in the input RGB or RGB-D videos. Taking inspiration from predecessors that leverage additional modalities for 3D/4D scene reconstruction~\cite{wilson2021echo, zhou2024drivinggaussian}, we propose to overcome this problem by fusing slow-speed RGB video with high-speed event streams.

\subsection{Event-based Vision}
Event cameras~\cite{lichtsteiner2008128,delbruck2010activity,gallego2020event} are neuromorphic sensors that record changes in illumination as discrete and asynchronous activation spikes, known as `events'. 
Compared to conventional frame-based cameras, event cameras offer advantages such as higher frame rate, higher dynamic range, and lower power, making them exceptionally suitable for high-speed and motion-sensitive tasks. 
Event signals are also insensitive to static backgrounds, which enables event-based visual computing algorithms to better concentrate on dynamic elements within a scene.

Due to the inherently high noise level of event streams, deep learning has been intensely used  to extract information from event cameras recordings. 
\emph{Event-based video reconstruction} aims to convert event streams into human-perceivable video frames. Representative works in this domain include E2VID~\cite{rebecq2019high}, FireNet~\cite{scheerlinck2020fast}, and Hypere2vid~\cite{ercan2024hypere2vid}. 
\emph{Event-based recognition and tracking} utilizes event cameras' high motion sensitivity to locate~\cite{gehrig2023recurrent}, segment~\cite{alonso2019ev, wan2025instance} and track~\cite{messikommer2023data} objects, with downstream applications such as eye-tracking~\cite{angelopoulos2021event} and egocentric action recognition~\cite{moreno2022visual}.
Finally, \emph{event-based motion estimation} regresses optical flows~\cite{gehrig2021raft, wan2022learning} or scene flow~\cite{ieng2017event} from event inputs.

\subsection{Event-assisted Scene Reconstruction}
Early efforts to incorporate event information into scene reconstruction primarily focused on static scenes; events were generated through camera motion. For instance, E2GS~\cite{deguchi2024e2gs} and Event3DGS~\cite{xiong2024event3dgs} conduct event-based deblurring of RGB frames, while EF-3DGS~\cite{liao2024ef} enhances camera pose estimation using event data. 
Simultaneously,  a line of research has been dedicated to scene reconstruction without the use of frame-based inputs. Noteworthy works in this domain include EvGGS~\cite{wang2024evggs}, EV-GS~\cite{wu2024ev}, Event-3DGS~\cite{han2024event}, Ev3DGS~\cite{huang2024ev3dgs} and SweepEvGS~\cite{wu2024sweepevgs}. These pioneering works, although not directly applicable to dynamic scenes, prove the feasibility of event-RGB fusion for scene reconstruction.

Only recently have event data been used in dynamic scene reconstruction. To date there have been two NeRF-based approaches, DE-NeRF~\cite{ma2023deformable} and Dynamic EventNeRF~\cite{rudnev2024dynamic}, and one GS-based method~\cite{xu2025event}. However, we note that all three  assume aligned event and RGB camera sensors. Furthermore, DE-NeRF and \cite{xu2025event} have only been validated on synthetic data or monocular real-world video. While Dynamic EventNeRF has been tested on realistic multiview event-RGB captures, the resolution and visual quality of their RGB inputs are extremely low and fail to match those of popular benchmarks such as Nvidia~\cite{yoon2020novel} and Neu3D~\cite{li2022neural}.

Although several large-scale event-RGB datasets, such as MVSEC~\cite{zhu2018multivehicle} and EventAid~\cite{duan2025eventaid}, have been published by the research community, they usually feature only one or two viewpoints,  insufficient for  scene reconstruction. To the best of our knowledge, no real-world multiview event-RGB dataset has been released, and therefore synthesizing event streams from multiview RGB videos using off-the-shelf simulators such as ESIM~\cite{rebecq2018esim} or v2e~\cite{hu2021v2e} is still the mainstream approach for validating such algorithms. 

In summary, the gap between existing methods and real-world scenarios stands out as the most crucial challenge for event-assisted dynamic scene reconstruction.

\section{Method}

Our proposed ERF-GS framework (see Fig.~\ref{fig:pipeline}) consists of an RGB-based dynamic GS backbone (see Sec.~\ref{sec:prelim}) and two components for fusing event information into the Gaussian representation, namely EARL (see Sec.~\ref{sec:earl}) and EDS (see Sec.~\ref{sec:eds}). Compared to other event-based GS methods, a notable distinction of ERF-GS is that neither EARL nor EDS requires ground truth RGB data in its computation. 
This design allows us to select any existing dynamic GS method as the backbone while upholding the model's robustness with insufficient and low-quality RGB video inputs.

\subsection{Preliminary: 3DGS for Dynamic Scenes}\label{sec:prelim}
As a learning-based solution to scene reconstruction, 3DGS represents the target scene as a collection of three-dimensional Gaussians. Each Gaussian, with  index $i$, is characterized by its center $\boldsymbol{\mu}_i$, covariance matrix $\boldsymbol{\Sigma}_i$, opacity $\sigma_i$, and spherical harmonic (SH) coefficients $\mathbf{h}_i$. Its visibility at spatial location $\mathbf{p}$ is defined as:
\begin{equation}
\label{eq:3dgs}
    \alpha_i = \sigma_i \exp\left[-\frac{1}{2}(\mathbf{p} - \boldsymbol{\mu}_i)^{\mathbf{T}}\Sigma_i^{-1}(\mathbf{p} - \boldsymbol{\mu}_i)\right].
\end{equation}

During the rendering process, the color at a given pixel  is determined by summing up the contributions of each Gaussian following a near-to-far order:
\begin{equation}
\label{eq:color}
    \mathbf{C} = \sum_{i=1}^N \mathbf{c}_i\alpha_i\prod_{j=1}^{i-1}(1 - \alpha_j),
\end{equation}
where $\mathbf{c}_i$ represents the RGB color of the $i^{\text{th}}$ Gaussian, and its visibility $\alpha_i$ is determined by using the 2D version of Eq.~\ref{eq:3dgs} and perspective projection. This differentiable rasterization pipeline enables fast rendering and the use of simple pixel-wise metrics such as the photometric $\ell_1$ loss to optimize the Gaussians' parameters.

To empower 3DGS with the capability to represent dynamic scenes, it is essential to model the temporal evolution of each 3D Gaussian. A representative approach is 4DGS~\cite{wu20244d}, which employs a deformation network $\phi$ to estimate the residual of Gaussian parameters at time $t$: $\phi (\boldsymbol{\mu}_0, t) \rightarrow \Delta_t (\boldsymbol{\mu}, \mathbf{r}, \mathbf{s}, \sigma, \mathbf{h})$, where $\mathbf{r}$ and $\mathbf{s}$ represent  rotation and scaling vectors, respectively. The parameters of $\phi$ are trained end-to-end along with all Gaussian attributes using the same losses as in 3DGS. 

In this paper, we use 4DGS as the backbone model for ERF-GS, yet our design is compatible with a wider range of dynamic 3DGS methods.

\subsection{Event-assisted Regularized Loss}\label{sec:earl}

Conventionally, an event stream is divided into multiple temporal bins and event activations within each bin are integrated into a 2D event frame for further processing. The event frame $E_{t\rightarrow t'}$ between $t$ and $t'$ characterizes the difference in log-intensity between intensity images. Formally:
\begin{equation}
    E_{t\rightarrow t'}^{(i,j)} = \frac{\ln \mathbf{I}_{t'}^{(i,j)} - \ln \mathbf{I}_{t}^{(i,j)}}{c},
\end{equation}
where $(i,j)$ indicates pixel coordinates (that we henceforth omit for conciseness) and $c$ is the contrast threshold. An event-based rendering loss can then be constructed as:
\begin{equation}
    \label{eq:old}
    \mathcal{L} = \left\|E_{t\rightarrow t'} - \frac{\ln R(\mathcal{G}, \rho, t') - \ln \mathbf{I}_{t}}{c}\right\|_1,
\end{equation}
where $R(\mathcal{G}, \rho, t')$ is the image rasterized from Gaussian representation $\mathcal{G}$ and camera pose $\rho$ at time $t'$. Despite being widely used by existing methods~\cite{xu2025event}, this formulation requires \emph{spatial, temporal, and channel-wise alignment} between event frames and RGB images, a challenging condition to attain in practice. Thus, we opt to substitute the ground truth RGB frame $\mathbf{I}_t$ in Eq.~\ref{eq:old} with another rendered image:
\begin{equation}
    \mathcal{L}_{\mathrm{event}} = \left\|E_{t\rightarrow t'} - \frac{\ln L[R(\mathcal{G}, \rho_e, t')] - \ln L[R(\mathcal{G}, \rho_e, t)]}{c}\right\|_1.
\end{equation}

By replacing $\rho$ with the event camera pose $\rho_e$, hereby omitted for conciseness, and applying the RGB-to-luma conversion $L(\cdot)$, we have lifted the dependency of the resulting loss term on ground truth RGB information and generalized it to monochrome event data from arbitrary viewpoints. However, supervising two rendered images simultaneously with the noisy and sparse $E_{t\rightarrow t'}$ results in suboptimal performance. Therefore, we enhance $\mathcal{L}_{\mathrm{event}}$ via the following weighting and regularization strategies.

\subsubsection{Confidence-weighed event loss}
The activation threshold $c$ determines the sensitivity of the event sensor to intensity changes, and in reality its value tends to fluctuate from event to event, which is the main cause of noise  in event data. Assuming that $c$ follows a normal distribution with variance $\sigma_c$, it is obvious that the actual log-intensity difference at a pixel conditioned on its event value $E$ is also distributed normally: $\Delta\ln\mathbf{I} \sim \mathcal{N}(cE, |E|\sigma_c)$. This observation suggests that we may formulate an event-based loss as the negative log likelihood (NLL) of the predicted event count given ground truth observation $E_{t\rightarrow t'}$:
\begin{equation}
    \mathcal{L}_{\mathrm{event}}^{\mathrm{cw}} = \left\|\frac{cE_{t\rightarrow t'} - \Big(\ln L[R(\mathcal{G}, t')] - \ln L[R(\mathcal{G}, t)]\Big)}{|E_{t\rightarrow t'}|}\right\|_2.
\end{equation}

Now, we are less confident in pixels with large event counts, as they accumulate larger errors from more event activations. $\mathcal{L}_{\mathrm{event}}^{\mathrm{cw}}$ is therefore a confidence-weighted version of $\mathcal{L}_{\mathrm{event}}$ that is more robust to `spikes' in event frames and is better suits scenes with large movements between two RGB frames.

\begin{figure}[t!]
  \centering
   \includegraphics[width=0.99\linewidth]{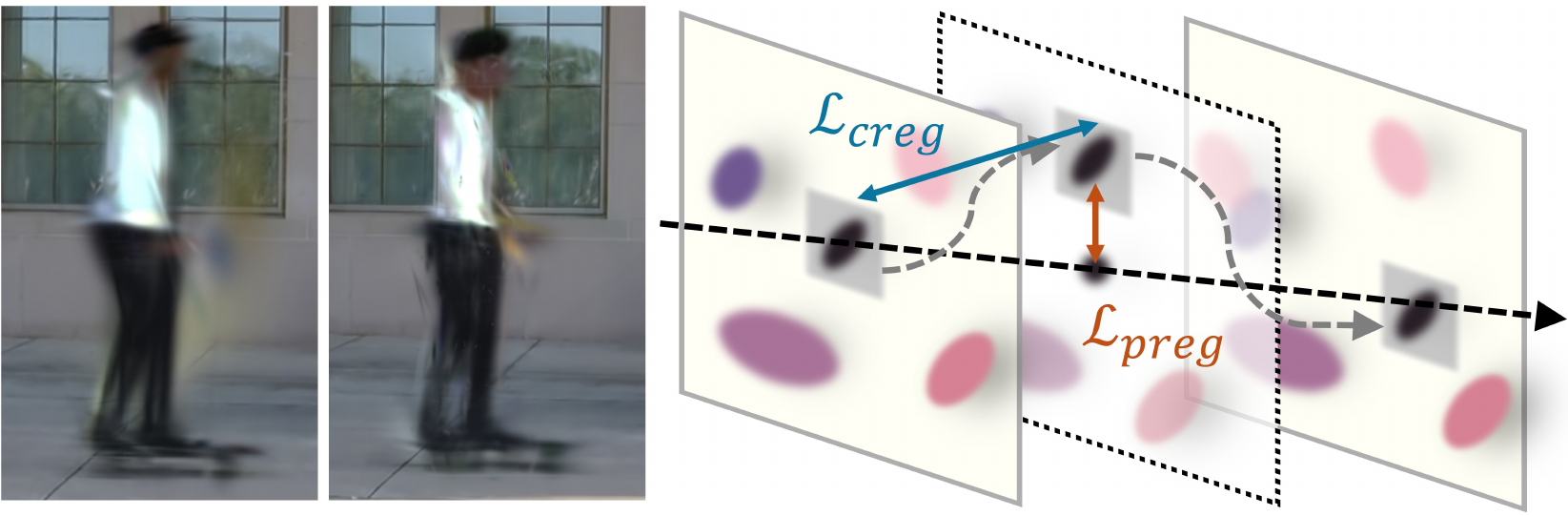}
   \caption{Left: training with monochrome event data can result in severe color shifts if not appropriately regularized. Right: the trajectory regularization loss $\mathcal{L}_\mathrm{preg}$ promotes linear movement of Gaussians between two RGB frames, while the color regularization loss $\mathcal{L}_\mathrm{creg}$ penalizes Gaussian color changes.}
   \label{fig:reg}
\end{figure}

\subsubsection{Trajectory and color regularization}
In  basic 4DGS, the Gaussian trajectories between two RGB frames are unconstrained. When trained with event-based loss functions, they tend to  zigzag to fit the sparse and noisy event data, greatly reducing the visual quality of the reconstructed video. Such behavior necessitates additional regularization on Gaussian trajectories.

For two consecutive RGB timesteps $t_0$ and $t_1$, we construct a reference position for each Gaussian at $t' \in (t_0, t_1)$ by linearly interpolating $\boldsymbol{\mu}_0$ and $\boldsymbol{\mu}_1$; they are relatively reliable since they are directly supervised by RGB frames. Then we penalize the deformed position $\hat{\boldsymbol{\mu}}$ at $t'$ as below:
\begin{equation}
    \mathcal{L}_{\mathrm{preg}} = \left\|\hat{\boldsymbol{\mu}} - \left[(\boldsymbol{\mu}_1 - \boldsymbol{\mu}_0)  \frac{t'- t_0}{t_1 - t_0} + \boldsymbol{\mu}_0\right]\right\|_1.
\end{equation}

\begin{figure*}[!t]
  \centering
   \includegraphics[width=\linewidth]{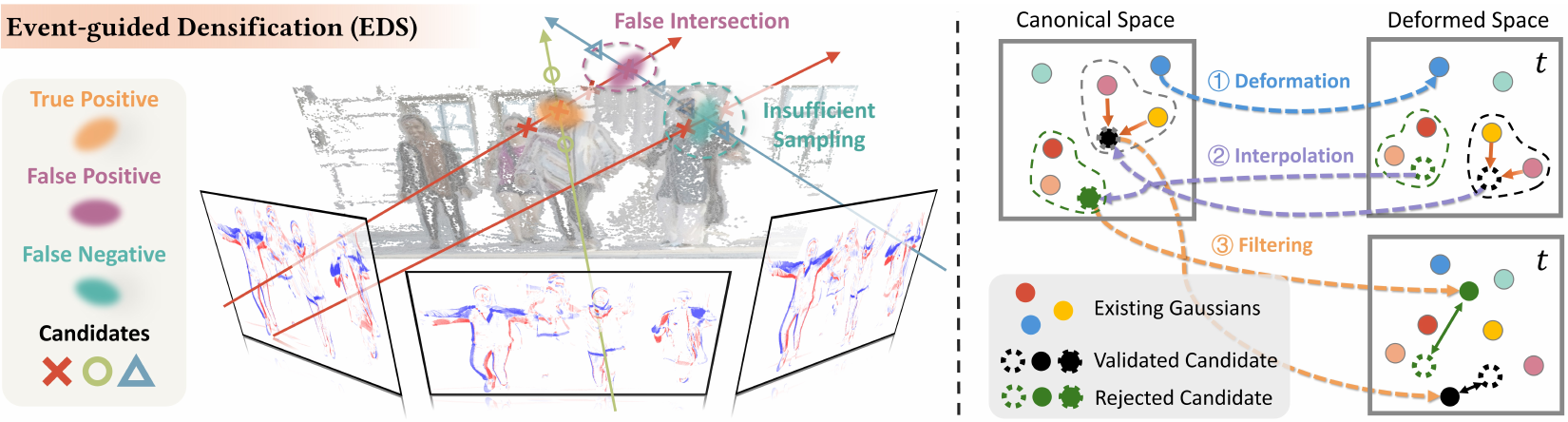}
   \caption{Left: we locate {Gaussian candidates} by sampling along rays cast from an event frame, and cross-validate them by comparison to other event viewpoints. However, interference to this  procedure arises from {false intersections} (where rays corresponding to different moving objects intersect in 3D space) and {insufficient sampling} (when intersection points fall between two candidates). Right: to embed selected candidates into the canonical space, we interpolate between their nearest neighbors in the deformed space, and further filter out those which deviate too much from their expected locations post-deformation.
   }
   \label{fig:eds}
\end{figure*}

In addition to trajectory instability, Gaussians supervised by monochrome events are also haunted by color shifts (see Fig.~\ref{fig:reg}(left)). Similarly to $\mathcal{L}_{\mathrm{preg}}$, we seek to regularize the color of each Gaussian at $t'$ to be close to that of $t_0$:
\begin{equation}
    \mathcal{L}_{\mathrm{creg}} = \|\hat{\mathbf{c}} - \mathbf{c}_0\|_1.
\end{equation}

The aims of the regularization terms are visualized in Fig.~\ref{fig:reg}(right). While other Gaussian parameters, including rotation, scaling, and opacity can also be regularized in a similar manner, we find empirically that doing so only marginally influences the reconstruction quality while introducing extra computation.

In summary, the event-assisted regularized loss (EARL) is the sum of all three loss terms, with weights $\lambda_p$ and $\lambda_c$:
\begin{equation}
    \mathcal{L}_{\mathrm{EARL}} = \mathcal{L}_{\mathrm{event}}^{\mathrm{cw}} + \lambda_p\cdot\mathcal{L}_{\mathrm{preg}} + \lambda_c\cdot\mathcal{L}_{\mathrm{creg}}.
\end{equation}

\subsection{Event-guided Densification}\label{sec:eds}

3DGS faces the challenge of insufficient Gaussians in complex regions, a problem that intensifies with object motion and blurred video input during fast dynamic scene modeling. Among existing solutions, gradient-based densification strategies~\cite{kerbl20233d} only incorporate Gaussians where the scene has already been represented, and unprojection-based algorithms~\cite{li2024spacetime} cannot accurately locate the depths of absent objects without consulting ground truth depth maps or pre-trained depth estimators. Thus, both methodologies lack the ability to populate unoccupied regions with new Gaussians when training dynamic GS models.

Fortunately, event data offers an accurate and efficient solution. Since event sensors are only activated by object motion, they naturally filter out static scene components and highlight moving regions for densification (assuming static camera poses). 
Moreover, the sparsity of event frames allows us to cross-validate candidate coordinates lifted from 2D frames with multiview references at a relatively low computation cost. Taking advantage of these properties, we propose an event-guided densification strategy (EDS), illustrated in Fig.~\ref{fig:eds}, that consists of three stages: unprojection, cross-validation, and filtering.

\subsubsection{Unprojection}
We generate 3D Gaussian candidates for densification by lifting pixels with nonzero event values from a given event frame (referred to as the ``source view'' hereafter) to 3D coordinates. 
For a pixel coordinate $p \in \mathbb{R}^2$ with $|E_p| > 0$, we firstly project a ray from the pixel location into the 3D space using the projection matrix $\mathbf{P}$. Subsequently, the resulting vector is normalized to  unit depth as follows:
\begin{align}
    \mathbf{v} &= \mathbf{P}^{-1}\begin{bmatrix}p_x \; p_y \; 1 \; 1\end{bmatrix}^{\mathbf{T}},\\
    \mathbf{v}_d &= \frac{\mathbf{v}}{\mathbf{v}_w\cdot\mathbf{v}_z}.
\end{align}

Following this, we sample $M$ depth values between $d_{\min}$ and $d_{\max}$, which are the minimum and maximum depths of the scene, respectively. We then construct candidate Gaussian positions as follows:
\begin{equation}
    \mathbf{p}_i = d_i \mathbf{v}_d,   
\end{equation}
where
\begin{equation}
  d_i = \frac{i}{M}(d_{\max} - d_{\min}) + d_{\min}.
\end{equation}

Finally, we can convert each $\mathbf{p}_i$ to the world frame using the camera-to-world transformation matrix $\mathbf{W}$: 
\begin{equation}
\hat{\mathbf{p}}_i = \mathbf{W}^{-1}\mathbf{p}_i.
\end{equation}

\subsubsection{Cross-validation}
The unprojected candidates are scattered evenly in  depth, and therefore, they cannot pinpoint the locations of moving objects. We  resolve this issue using multiview cross-validation. Specifically, we select $K$ event frames from the same timestep as reference views and use their camera matrices $\mathbf{W}_k$ and $\mathbf{P}_k$ to re-project each candidate's $\hat{\mathbf{p}}_i$ to 2D:
\begin{equation}
    p_i^{k} = \mathbf{P}_k\mathbf{W}_k\hat{\mathbf{p}}_i.
\end{equation}

We only retain $\hat{\mathbf{p}}_i$ if its projected coordinates $p_i^k$ have event values for all $K$ reference views. We find that a relatively small $K \ge 2$ suffices to perform effective cross-validation and yield satisfactory densification results.

\subsubsection{Filtering}
Even after cross-validation, the resulting candidates $\mathcal{P}$ are still unreliable due to \emph{false intersections} and \emph{insufficient sampling} (see Fig.~\ref{fig:eds}(left)). False intersection occurs when a candidate projects to different scene contents in different views, while insufficient sampling happens when an intersection fails cross-validation because it happens to lie between two adjacent samples. These phenomena occur more frequently when the number of reference views is limited. We employ three steps to remove outliers from $\mathcal{P}$: depth-wise uniqueness filtering, spatial coherence filtering, and deformation consistency filtering. 

\paragraph{Depth-wise uniqueness filtering}
During the unprojection stage, multiple candidates with varying depths are generated for each pixel, while only those visible to the camera contribute to producing event activations. 
Therefore, we filter $\mathcal{P}$ to keep only the candidate with the smallest depth for each ray cast from the source view.

\paragraph{Spatial coherence filtering}
We  filter out candidates that are distant from established Gaussians, as they are more likely to arise from noise events. As events can be triggered by both the appearance and disappearance of objects, we compute the deformed positions of existing Gaussians at two adjacent timesteps $t_0 < t < t_1$, denoted  $\mathcal{P}_0$ and $\mathcal{P}_1$, respectively. Utilizing the $k$-nearest neighbors (kNN) algorithm, we calculate the spatial coherence of the $i^{\text{th}}$ candidate to be:
\begin{equation}
    C_i = \min\left(\frac{1}{\kappa}\sum_{j\in R_0} \left\|\hat{\mathbf{p}}_i - \mathcal{P}_0^{(j)}\right\|_2,\  \frac{1}{\kappa}\sum_{j\in R_1} \left\|\hat{\mathbf{p}}_i - \mathcal{P}_1^{(j)}\right\|_2\right)
\end{equation}
where $R_0$ and $R_1$ are the set of $\kappa$-nearest neighbors in $\mathcal{P}_0$ and $\mathcal{P}_1$. Subsequently, we randomly sample $N$ candidates from the top $2N$ with the smallest $C_i$ and discard the remainder.

\paragraph{Deformation consistency filtering}
In deformation network-based methods such as 4DGS~\cite{wu20244d}, Gaussian parameters are stored in a canonical space and a non-invertible neural network $\phi$ is employed to deform them at any given timestep. Therefore, the canonical coordinates of densification candidates $\mathcal{P}$ can only be backward localized from their nearest neighbors. Formally:
\begin{equation}
\label{eq:knn}
    \hat{\boldsymbol{\mu}}_0^i = \frac{1}{\kappa} \sum_{j\in R} \boldsymbol{\mu}_0^j,\ R = \mathrm{kNN}(\hat{\mathbf{p}}_i, \mathcal{P}_t, \kappa),
\end{equation}
where $\mathcal{P}_t$ represents the deformed centers of existing Gaussians at $t$. However, the behavior of $\phi$ over the whole 3D space is unconstrained and the actual $\phi(\hat{\boldsymbol{\mu}}_0^i)$ post deformation could deviate from the intended position $\hat{\mathbf{p}}_i$. Our ultimate filtering step seeks to remove candidates with deviation larger than a given threshold $\epsilon$. Figure~\ref{fig:eds}(right) demonstrates the backward localization and deformation consistency filtering process.
Note that for dynamic GS methods utilizing invertible deformation functions or frame-wise Gaussian parameters, this filtering step may be omitted. 
Other parameters for newly added Gaussians, such as rotation, scale, and opacity, can also be interpolated from their nearest neighbors in a way similar to Eq.~\ref{eq:knn}.

\section{Experiments}\label{sec:exp}

\subsection{Datasets}
\label{sec:exp1}
Creating multiview event-RGB video datasets poses several engineering challenges and demands substantial investments in cost and time.  
Existing methods~\cite{mueggler2017event} either employ 3D graphics software such as Blender, combined with event simulation algorithms, to render high speed event-RGB videos, or directly utilize pre-existing event-RGB video datasets for scene reconstruction. 
However, the former approach lacks realism in color and texture, while the latter is constrained by a limited range of viewpoints. Notably, neither approach serves as a robust benchmark for evaluating 3D scene reconstruction in real-world settings.

While we have not gone as far as collecting and experimenting with real-world multiview event-RGB data in this work, we still aimed to increasing the realism of simulated datasets to better evaluate our proposed ERF-GS framework.
Thus, we used the following pipeline to simulate both events and fast object motion for existing multiview natural video. 
Firstly, we applied v2e~\cite{hu2021v2e}, a state-of-the-art event simulator, to each RGB video to generate synthetic events with realistic noise distribution. 
Subsequently, we implemented three steps to progressively align our datasets with real-world scenarios featuring fast scene dynamics (or equivalently, low frame rates) and disjoint event-RGB camera configurations. 
The first step involves temporally subsampling the RGB video by a factor  $f \in \{4, 6, 8\}$ based on its original duration.

\begin{figure}[t!]
  \centering
   \includegraphics[width=\linewidth]{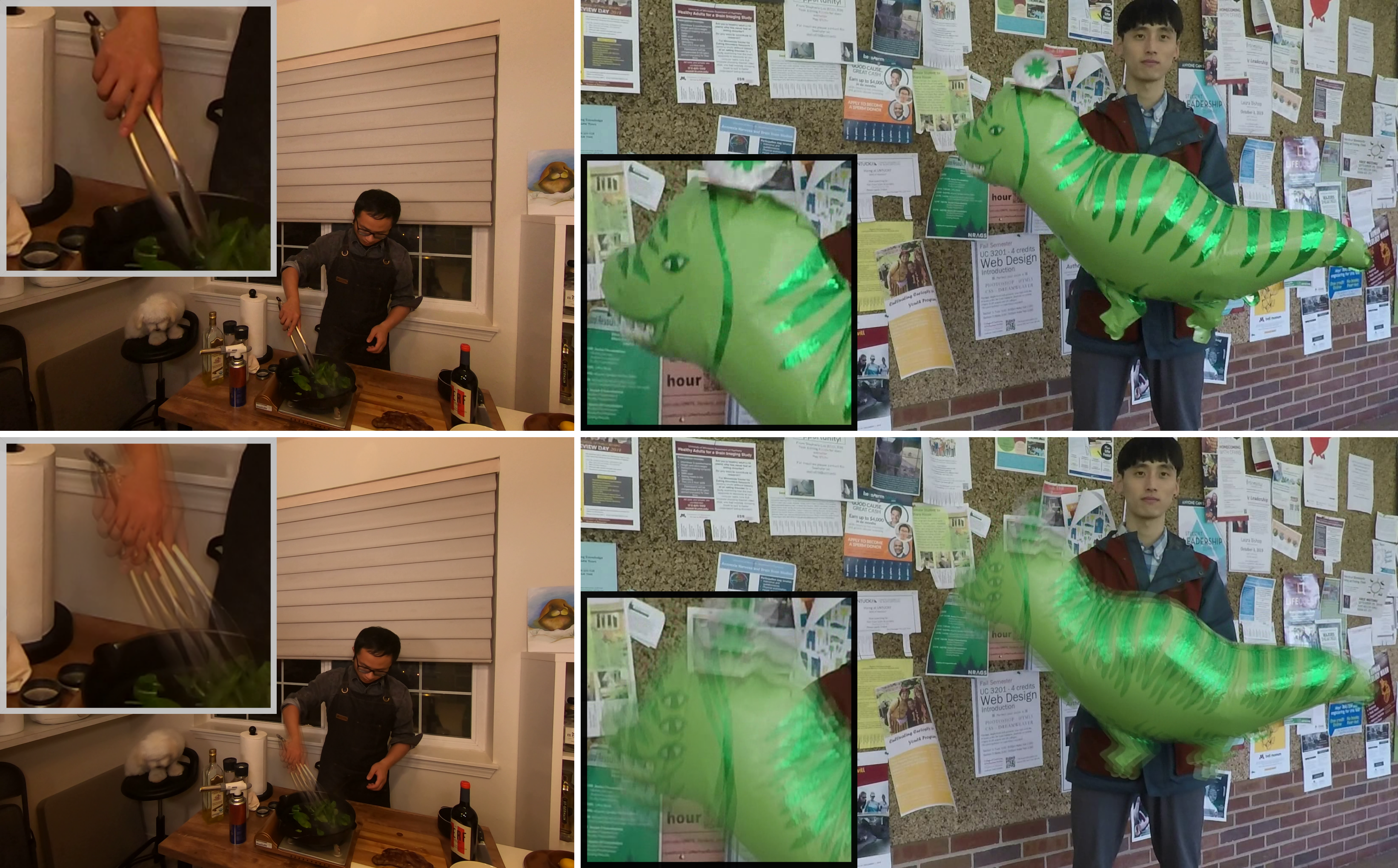}
   \caption{Sample RGB frames before (above) and after (below) applying simulated motion blur. Both the visual details and the spatial positions of moving objects are corrupted by this operation.
   }
   \label{fig:mbdv}
\end{figure}

Secondly, we simulate motion blur by averaging each frame after subsampling with its two neighbors from the original video, as shown in Fig.~\ref{fig:mbdv}.
Although it would be possible to apply more realistic motion blur simulations (e.g. by estimating optical flow and constructing blur kernels for each pixel), we find that such methods break the visual consistency between different views and largely degrade the performance of GS-based reconstruction.

Lastly, we partition the training set of viewpoints $\mathcal{V}$ into separate sets, $\mathcal{V}_{\mathrm{RGB}}$ and $\mathcal{V}_{\mathrm{event}}$, ensuring that event streams and RGB videos are captured with different camera poses. 
For clarity, we mark the datasets after each step with suffixes {-ts}, {-mb} and {-dv} (short for {t}emporal {s}ubsampling, {m}otion {b}lur and {d}isjoint {v}iews, respectively). We select 2--3 views from each scene without any transformation as the test set in order to evaluate the ability of each model to interpolate between sparse RGB frames and its robustness to motion blur.

In the hope of mitigating the lack of standard benchmarks in this field, we base our datasets on two popular dynamic reconstruction benchmarks, Neu3D~\cite{li2022neural} and Nvidia~\cite{yoon2020novel}. Neu3D consists of six dynamic scenes of a person cooking; each scene has 18--20 multiview videos with 2K resolution and 300 frames. Nvidia is a more challenging dataset featuring eight indoor and outdoor scenes with complex dynamics; each scene has 12 multiview videos with 2K resolution and 90--200 frames. To generate their {-ts}, {-mb} and {-dv} variants, we used the following configurations: The temporal subsampling rate $f$ was set to 16 across all scenes in Neu3D. In Nvidia, $f$ was chosen to be 4 for `Jumping', 6 for `Playground' and `Skating', and 8 for all  remaining scenes. After subsampling, each scene was characterized by 18--24 RGB frames per view. The scene `Balloon2' was excluded from the Nvidia dataset due to consistently poor reconstruction quality for all tested methods. For Nvidia-dv, 3 uniformly sampled views from each scene were designated as event-only views, while for Neu3D-dv, 4 views were uniformly sampled for this purpose.

Note that although we experimented exclusively on datasets with static cameras, ERF-GS does not assume pose invariance across frames and therefore it can be readily applied to scenes with moving cameras without changing its core modules.

\subsection{Implementation and training}
\subsubsection{General}
We initialized the Gaussian point cloud and obtained camera poses by running COLMAP~\cite{schoenberger2016sfm}  on RGB frames for each scene. Our ERF-GS models were based on and compared to the 4DGS~\cite{wu20244d} backbone. 
For the Neu3D dataset variants, we mirrored the configuration of 4DGS and trained the model for 3,000 coarse iterations without employing the deformation network, followed by 14,000 fine iterations. 
For the Nvidia dataset variants, we increased the depth of deformation network to 2 and doubled the network width to 256 to accommodate the increased scene complexity. The model was trained for 3,000 coarse iterations and 30,000 fine iterations. 
All training was done on an nVidia GeForce RTX 4090 GPU.

\begin{table*}[t!]
\caption{{Quantitative results on Neu3D-dv}. Metrics reported are: DPSNR (dB)$\uparrow$, PSNR (dB)$\uparrow$, SSIM$\uparrow$ and LPIPS$\downarrow$.}
\centering
{
\begin{tabular}{l|cc}
\toprule
Scene & \textbf{4DGS} & \textbf{ERF-GS (Ours)} \\ 
\midrule
coffee\_martini & 25.86 / 24.12 / 0.866 / 0.193 & \cellcolor{best}\textbf{26.78} / 24.13 / 0.868 / 0.185\\
cook\_spinach & 25.63 / 28.40 / 0.927 / 0.156 & \cellcolor{best}\textbf{27.05} / 28.70 / 0.930 / 0.155\\
cut\_roasted\_beef & \cellcolor{best}\textbf{27.17} / 27.35 / 0.914 / 0.167 & 26.09 / 27.16 / 0.908 / 0.165\\
flame\_salmon & 22.91 / 23.21 / 0.865 / 0.187 & \cellcolor{best}\textbf{24.71} / 23.35 / 0.869 / 0.179\\
flame\_steak & 25.18 / 27.72 / 0.927 / 0.153 & \cellcolor{best}\textbf{26.99} / 28.16 / 0.931 / 0.144\\
sear\_steak & 27.69 / 25.91 / 0.900 / 0.168 & \cellcolor{best}\textbf{28.26} / 26.26 / 0.901 / 0.166\\
\midrule 
\textbf{Avg.} & 25.74 / 26.12 / 0.900 / 0.171 & \cellcolor{best} \textbf{26.65} / 26.29 / 0.901 / 0.166 \\
\bottomrule
\end{tabular}
}
\label{tab:neu3d-mbdv}
\end{table*}

\begin{table*}[t!]
\caption{{Quantitative results on Nvidia-dv}. Metrics reported are: DPSNR (dB)$\uparrow$, PSNR (dB)$\uparrow$, SSIM$\uparrow$ and LPIPS$\downarrow$.}
\centering
{
\begin{tabular}{l|cc}
\toprule
Scene & \textbf{4DGS} & \textbf{ERF-GS (Ours)} \\ 
\midrule
Balloon1 & 20.10 / 21.81 / 0.704 / 0.296 & \cellcolor{best}\textbf{21.84} / 22.47 / 0.704 / 0.287\\
Skating & 14.51 / 26.51 / 0.878 / 0.194 & \cellcolor{best}\textbf{17.28} / 27.54 / 0.880 / 0.195\\
Dynamicface & \cellcolor{best}\textbf{22.81} / 17.18 / 0.722 / 0.256 & 22.63 / 16.83 / 0.697 / 0.301\\
Jumping & 16.95 / 23.11 / 0.810 / 0.254 & \cellcolor{best}\textbf{18.58} / 24.17 / 0.817 / 0.254\\
Truck & 21.83 / 23.70 / 0.769 / 0.252 & \cellcolor{best}\textbf{22.43} / 24.44 / 0.799 / 0.261\\
Playground & 17.43 / 20.78 / 0.667 / 0.290 & \cellcolor{best}\textbf{19.35} / 20.81 / 0.670 / 0.278\\
Umbrella & 19.00 / 22.29 / 0.561 / 0.356 & \cellcolor{best}\textbf{20.60} / 23.13 / 0.565 / 0.334\\
\midrule 
\textbf{Avg.} & 18.95 / 22.20 / 0.730 / 0.271 & \cellcolor{best}\textbf{20.39} / 22.77 / 0.733 / 0.273\\
\bottomrule
\end{tabular}
}
\label{tab:nvidia-mbdv}
\end{table*}

\subsubsection{EARL}
The weight for $\mathcal{L}_\mathrm{event}$ was set to 0.1, while the  regularization weights $\lambda_p$ and $\lambda_c$ were both set to 0.01 across all our experiments. As in the 4DGS method, we adopt  $\ell_1$ loss as the sole photometric loss. 
As event values are integers, we rounded  the predicted event $E'$ and applied the straight-through estimator (STE) to facilitate  backward propagation:
\begin{align}
    \hat{E}' = 
\begin{cases}
    \lfloor E' \rfloor,& \text{if } E' \ge 0,\\
    \lceil E' \rceil,              & \text{otherwise}.
\end{cases}\\
E_{\mathrm{STE}}' = E' + \mathrm{sg}(\hat{E}' - E').
\end{align}

\subsubsection{EDS}
Important hyperparameters set for EDS include: the number of candidates on each ray $M = 100$, the threshold in deformation consistency filtering $\epsilon = 10$, the number of nearest neighbors $\kappa = 3$, and the number of candidates for each densification step $N = 500$. To streamline computation, we downsampled input event frames to $0.25\times 0.25$ their original resolution. We ran  EDS every 200 iterations, starting from the $500^\mathrm{th}$ fine iteration.

\subsection{Results}

\begin{figure}[t!]
  \centering
     \includegraphics[width=\linewidth]{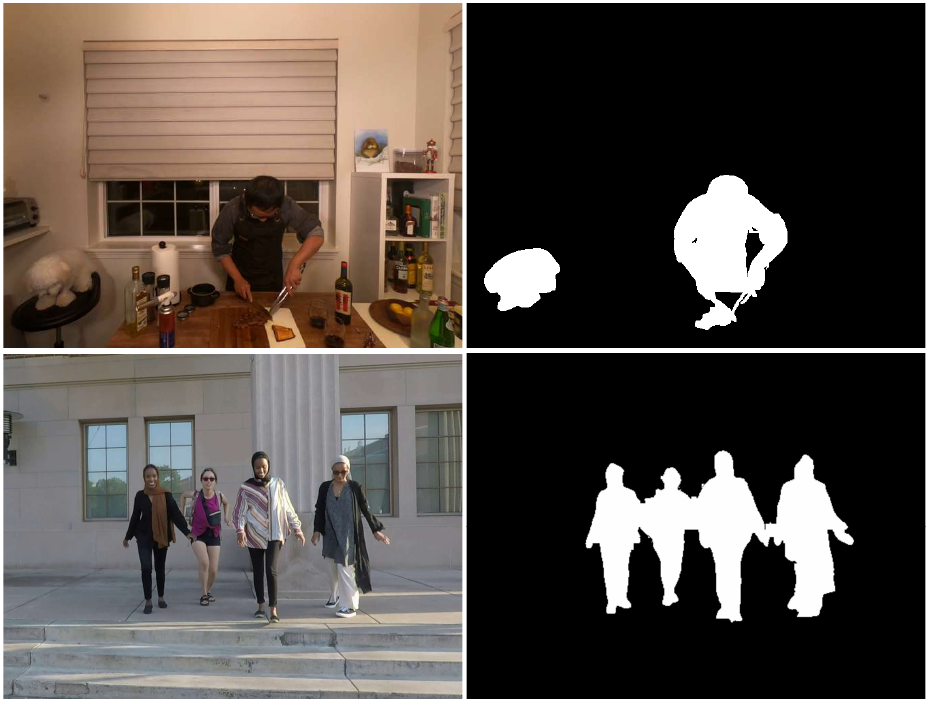}
   \caption{Two examples of SAM v2-generated dynamic masks: `cut\_roasted\_beef' (above) and `Jumping' (below).
   }
   \label{fig:mask}
\end{figure}

Quantitative results on Neu3D-dv and Nvidia-dv are presented in Tables~\ref{tab:neu3d-mbdv} and~\ref{tab:nvidia-mbdv}, respectively. In addition to conventional metrics such as PSNR, SSIM, and LPIPS, we report dynamic PSNR (DPSNR) by segmenting each video with SAM v2~\cite{ravi2024sam2} (see Fig.~\ref{fig:mask} for examples) and subsequently computing the PSNR on moving pixels only. In this way, the dynamic modeling ability of each method on scenes with large static backgrounds and small moving objects can be more accurately evaluated. 
Our findings indicate that ERF-GS outperforms the 4DGS baseline in learning fast dynamic scenes, with  0.17~dB and 0.57~dB gains in PSNR on Neu3D-dv and Nvidia-dv, respectively. 

\begin{table}[t]
\caption{{Quantitative results on different variants of two base datasets}. DPSNR$\uparrow$ and PSNR$\uparrow$ are reported.}
\centering
{
\setlength\tabcolsep{0pt}
\begin{tabular*}{\columnwidth}{@{\extracolsep{\fill}} lccc}
\toprule
Dataset & \textbf{4DGS} & \textbf{ERF-GS (Ours)} & \textbf{Relative gain} \\
\midrule
Nvidia-ts & 19.08 / 22.63 & \cellcolor{best}\textbf{21.62} / \textbf{23.78} & +2.54 / +1.15\\
Nvidia-mb & 19.15 / 22.71 & \cellcolor{best}\textbf{20.99} / \textbf{23.32} & +1.84 / +0.61\\
Nvidia-dv & 18.95 / 22.20 & \cellcolor{best}\textbf{20.39} / \textbf{22.77} & +1.44 / +0.57\\
\midrule
Neu3D-ts & 25.66 / 27.20 & \cellcolor{best}\textbf{27.51} / \textbf{27.49} & +1.85 / +0.29\\
Neu3D-mb & 25.56 / 27.03 & \cellcolor{best}\textbf{27.77} / \textbf{27.59} & +2.21 / +0.56\\
Neu3D-dv & 25.74 / 26.12 & \cellcolor{best}\textbf{26.65} / \textbf{26.29} & +0.91 / +0.17\\
\bottomrule
\end{tabular*}
}
\label{tab:tsmb}
\end{table}

\begin{figure*}[!tp]
  \centering
     \includegraphics[width=\linewidth]{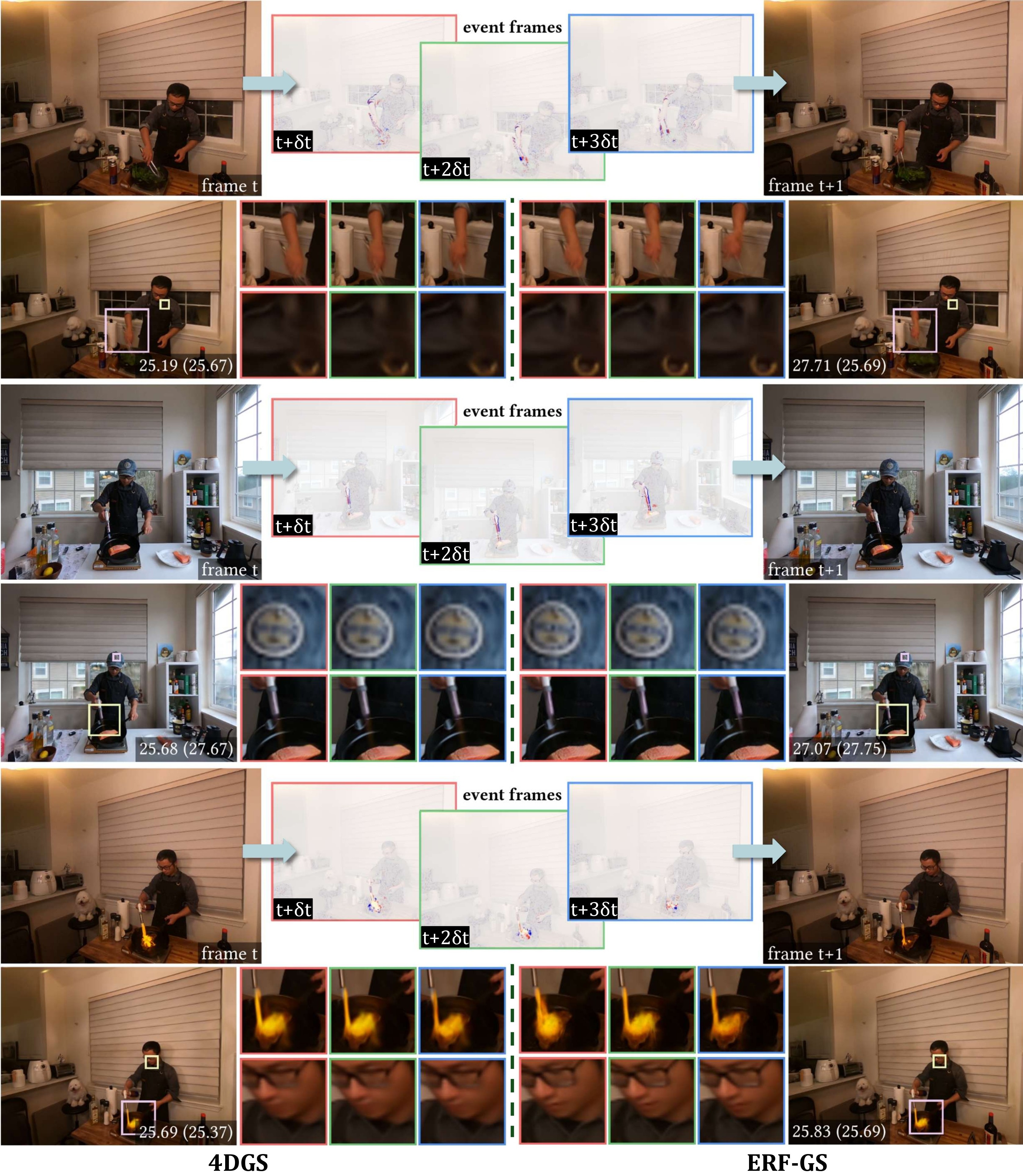}
   \caption{{Qualitative comparison on Neu3D-dv,}  showing three scenes: `cook\_spinach', `flame\_salmon', and `flame\_steak'. For each scene, the top row shows two consecutive RGB frames and three intermediate event frames, while the bottom row shows images reconstructed  with the 4DGS baseline (left) and ERF-GS (right). Patches are extracted at timesteps corresponding to the event frames, distinguished by bounding box colors. Metrics reported are DPSNR$\uparrow$ (PSNR$\uparrow$) in dB. For clarity we use the same viewpoint for all visualizations within a scene, even though in reality they have different camera poses.
   }
   \label{fig:large-neu3d}
\end{figure*}

\begin{figure*}[!tp]
  \centering
   \includegraphics[width=\linewidth]{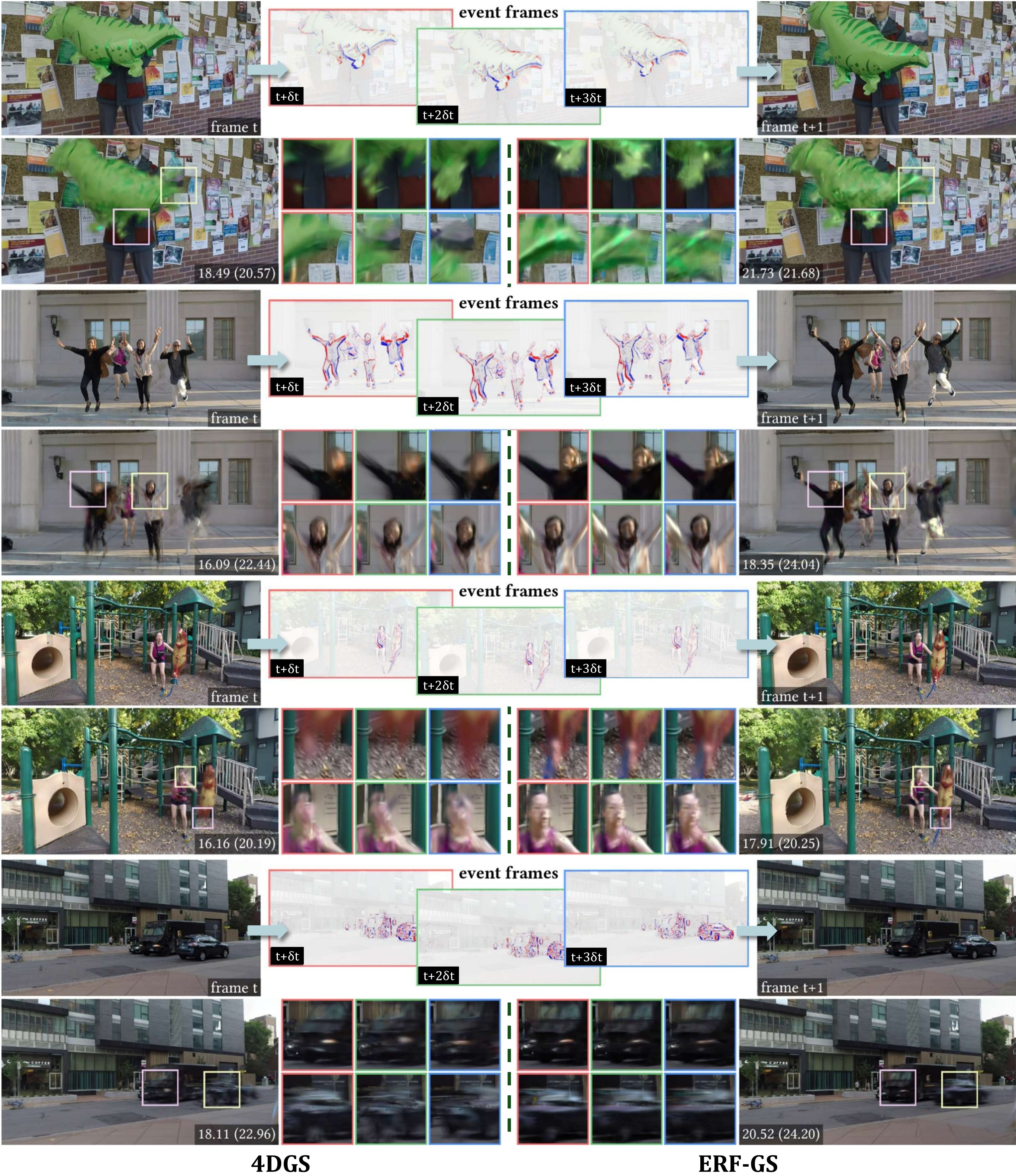}
   \caption{{Qualitative comparison on Nvidia-dv,} showing four scenes: `Balloon1', `Jumping', `Playground', and `Truck', are visualized. Layout and meaning is as in  Fig.~\ref{fig:large-neu3d}.
   }
   \label{fig:large-nvidia}
\end{figure*}

A more indicative metric is DPSNR, since event-based supervision is only effective for scene moving content, and our method shows a significant increase of 0.91~dB and 1.44~dB, respectively, validating its claimed effectiveness for modeling fast object motion. We also present  qualitative comparisons in Figs.~\ref{fig:large-neu3d} and~\ref{fig:large-nvidia}. One can straightforwardly observe that ERF-GS not only recovers better details of dynamic scenes (e.g., human faces in `Playground'), but also more faithfully follows the ground truth motion of objects (e.g., dinosaur feet in `Balloon1' and hand movement in `cook\_spinach').
\FloatBarrier

\begin{figure}[t]
  \centering
     \includegraphics[width=\linewidth]{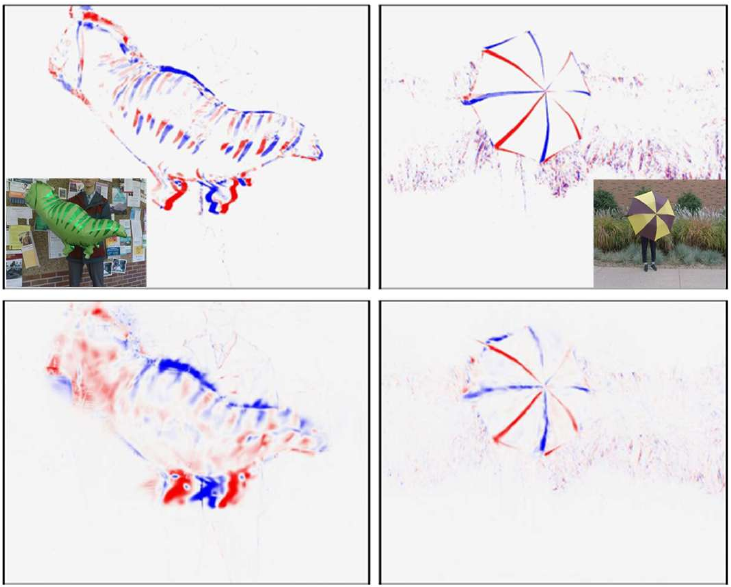}
   \caption{{Qualitative comparison on event frames.} Above: simulated event frames from `Balloon1' and `Umbrella', and their corresponding RGB frames.  Below: estimated event frame using two rendered images, exhibiting both successful reproduction of edges and textures (stripes on the balloon and umbrella ridges) and failures (head of balloon).
   }
   \label{fig:supp2}
\end{figure}

We further provide a  qualitative comparison of simulated event frames and those estimated from two rendered frames in Fig.~\ref{fig:supp2} to assess the dynamic modeling capability of ERF-GS in the event domain. Two scenes were chosen featuring distinct dynamic patterns: large rigid translation (`Balloon1' from Nvidia-dv) and rotation (`Umbrella' from Nvidia-dv). It can be observed that ERF-GS learns both types of motion with good quality. However, it may still fail with overly complicated textures such as the head of the dinosaur balloon.

In addition to the most challenging Nvidia-dv and Neu3D-dv datasets, we  evaluate the performance of ERF-GS on their {-mb} and {-ts} variants in Table~\ref{tab:tsmb}. In those experimental settings where event and RGB inputs share the same camera poses, our method takes advantage of the increased diversity of event viewpoints and the synchronicity between the two modalities to establish an even larger margin over the baseline ($\approx2$~dB for both datasets in DPSNR), both with and without motion blur in RGB videos. 

While it may appear counter-intuitive that model performance on {-mb} variants does not necessary fall behind {-ts} variants (\emph{e.g.}, 4DGS on Nvidia and ERF-GS on Neu3D), we attribute this to the fact that simulated motion blur carries information about adjacent frames, which themselves are excluded from training due to temporal subsampling. Consequently, reconstructed scenes may exhibit smoother motion at test timestamps when trained with {-mb} variants. Moreover, motion blur corrupts object textures and makes model convergence easier, as can also be observed from the larger training PSNR/SSIM of {-mb} variants in Table~\ref{tab:trainset}.

\begin{table}[t]
\caption{{PSNR$\uparrow$ and SSIM$\uparrow$ on training sets}. Even at time steps where the models are supervised by ground truth RGB frames, ERF-GS exhibits a small yet steady advantage over the baseline.}
\centering
{
\begin{tabular}{l|cc}
\toprule
Dataset & \textbf{4DGS} & \textbf{ERF-GS (Ours)}\\ 
\midrule
Nvidia-dv & 28.09 / 0.833 & \cellcolor{best}\textbf{28.37} / 0.842\\
Neu3D-dv & 33.93 / 0.945 & \cellcolor{best}\textbf{34.66} / 0.956\\
\midrule
Nvidia-mb & \cellcolor{best}\textbf{29.40} / 0.866 & 29.19 / 0.863\\
Neu3D-mb & 36.42 / 0.966 & \cellcolor{best}\textbf{36.45} / 0.966\\
\midrule
Nvidia-ts & 28.32 / 0.855 & \cellcolor{best}\textbf{28.43} / 0.854\\
Neu3D-ts & 35.58 / 0.958 & \cellcolor{best}\textbf{35.65} / 0.958\\
\bottomrule
\end{tabular}
}
\label{tab:trainset}
\end{table}

Note that the reconstruction quality of ERF-GS on {-dv} variants not only surpasses 4DGS in the same setting where the baseline sees fewer viewpoints (since $|\mathcal{V}_\mathrm{event}|$ cameras are selected as `event only'), it also outperforms the baseline model trained on {-mb} datasets, which utilizes the full set of RGB viewpoints, by 1.24~dB and 1.09~dB in DPSNR for Nvidia and Neu3D, respectively. This comparison shows that the advantage of ERF-GS on Neu3D-dv and Nvidia-dv does not result from merely seeing more viewpoints, but rather from its ability to leverage the high temporal resolution information in event inputs.

We also compare the PSNR and SSIM of the 4DGS baseline and our ERF-GS on the training set of each dataset (that is, on time steps where ground truth RGB images are used to train the models) in Table~\ref{tab:trainset}.

\begin{table*}[!t]
\caption{{Ablation study for ERF-GS.} DPSNR$\uparrow$ and PSNR$\uparrow$ are reported.}
\centering
{
\setlength\tabcolsep{0pt}
\begin{tabular*}{\linewidth}{@{\extracolsep{\fill}} lcc}
\toprule
Method & \textbf{Skating (Nvidia-dv)} & \textbf{Truck (Nvidia-dv)} \\ 
\midrule
4DGS baseline & 14.51 / 26.51 & 21.83 / 23.70\\
w/ EARL only & 17.10 / 27.58 & 22.17 / 23.85\\
w/ EDS only & 14.62 / 26.61 & 22.00 / 23.77\\
ERF-GS, no regularization & 17.27 / 27.34 & 22.38 / 23.91\\
\midrule 
\textbf{ERF-GS} & \textbf{17.28} / 27.54 & \textbf{22.43} / 24.44\\
\bottomrule
\end{tabular*}
}
\label{tab:ablation}
\end{table*}

\subsection{Ablation and Related Studies}

\begin{figure}[!t]
  \centering
   \includegraphics[width=\linewidth]{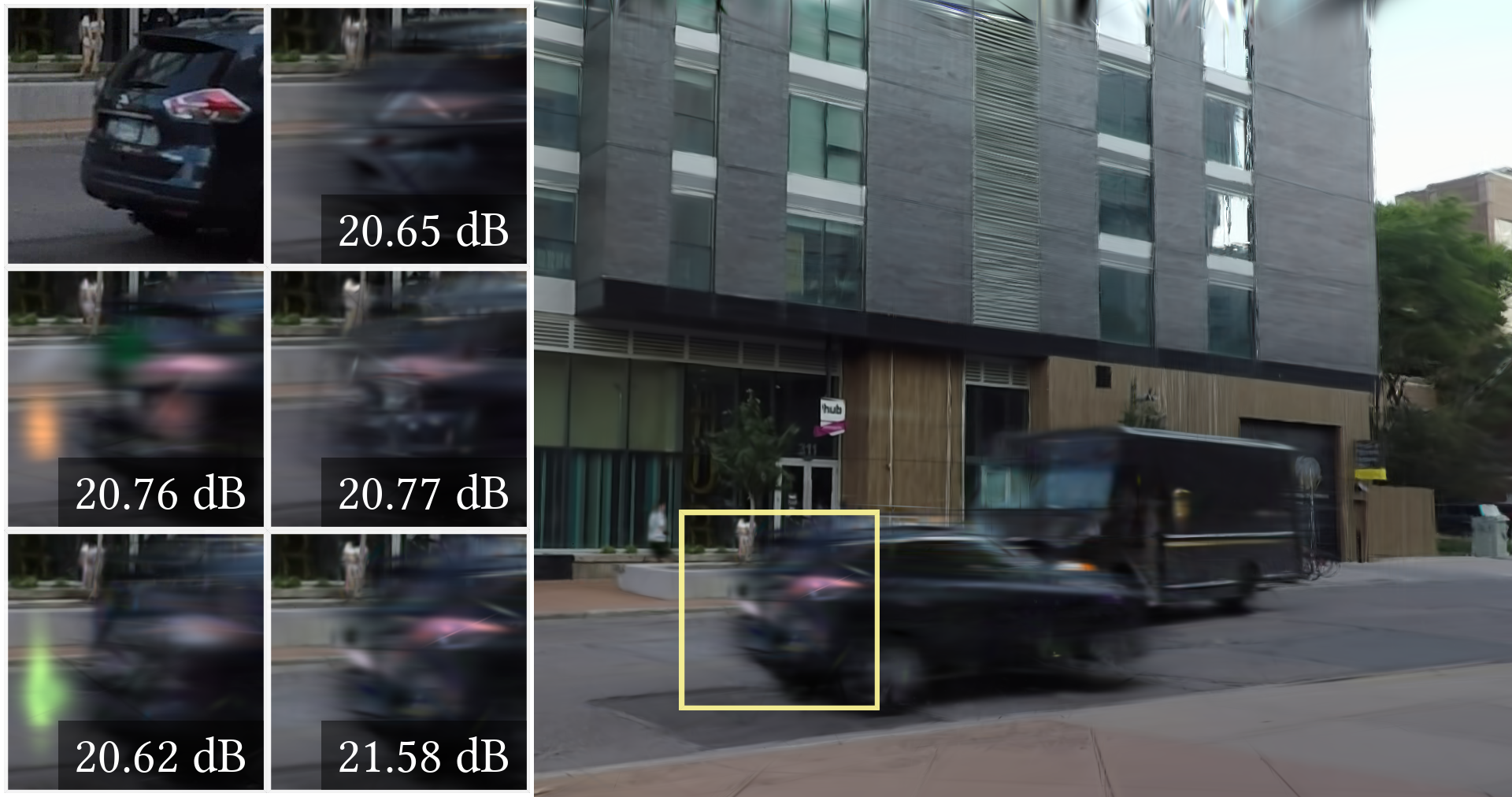}
   \caption{{Ablation study using `Truck' from Nvidia-dv.} Left to right: top row: ground truth, 4DGS baseline, middle row: with EARL only, with EDS only, bottom row: ERF-GS without regularization terms, ERF-GS. PSNR$\uparrow$ of each patch is provided for reference. 
   }
   \label{fig:ablation}
\end{figure}

We have conducted an ablation study on two scenes from Nvidia-dv, `Skating' and `Truck', which demonstrate challenging cross-screen object translation and appearance/vanishing of dynamic objects, making the advantages of EARL and EDS more significant. The aim was to demonstrate the effects of EARL, EDS, and the regularization losses on reconstructed dynamic scenes. From the results reported in Table~\ref{tab:ablation}, it can be observed that each proposed component contributes to the superior reconstruction power of ERF-GS, especially around dynamic regions. A close examination of the reconstructed scenes in Fig.~\ref{fig:ablation} also validates the effectiveness of EARL and EDS, and that the usage of regularization does reduce aberrant colors that linger behind the moving vehicle.

We also briefly investigated how the number of disjoint event viewpoints (\#DV) influences the performance of ERF-GS. As Table~\ref{tab:ablation2} shows, both DPSNR and PSNR remain relatively stable for different choices of \#DV, demonstrating the robustness of ERF-GS to varying capture setups.

\begin{table}[!t]
\footnotesize
\caption{{Study varying the number of disjoint event viewpoints.} DPSNR$\uparrow$ and PSNR$\uparrow$ are reported.}
\centering
{
\begin{tabular}{l|cc}
\toprule
\#DV & \textbf{Skating (Nvidia-dv)} & \textbf{Truck (Nvidia-dv)} \\ 
\midrule
2 & 17.21 / 27.91 & 22.42 / 24.43\\
3 (our method) & 17.28 / 27.54 & 22.43 / 24.44\\
4 & 17.07 / 26.64 & 22.47 / 24.17\\
\bottomrule
\end{tabular}
}
\label{tab:ablation2}
\end{table}

Introducing additional modules is expected to increase the overall training time; we provide a quantitative comparison in Table~\ref{tab:time} to better characterize the tradeoff between efficiency and accuracy. Note that EARL accounts for most of the increased training time in ERF-GS. Although it seems counterintuitive that applying EDS results in faster model training for the scene `Balloon1', such behavior can in fact be attributed to the smaller Gaussian point cloud size in the configuration `4DGS + EDS', since EDS directly adds new Gaussians and may affect the original densification and pruning process. Since we do not make any change to the inference pipeline of 4DGS, ERF-GS maintains a relative fast inference speed: ~120 fps on Neu3D and ~50 fps on Nvidia, respectively.

\begin{table}[!t]
\caption{{Training times for different models.} Relative training times (as percentages) in comparison to the 4DGS baseline are reported.}
\centering
{
\begin{tabular}{l|cc}
\toprule
Method & \textbf{Neu3D-dv } & \textbf{Nvidia-dv}\\
 & \textbf{(coffee\_martini)} & \textbf{(Balloon1)} \\
\midrule
4DGS + EARL  & +10.1\% & +25.8\%\\
4DGS + EDS & +3.2\% & -7.3\%\\
ERF-GS & +14.0\% & +28.8\%\\
\bottomrule
\end{tabular}
}
\label{tab:time}
\end{table}

\subsection{Comparison to Event-based Baseline}

\begin{figure}[t]
  \centering
   \includegraphics[width=\linewidth]{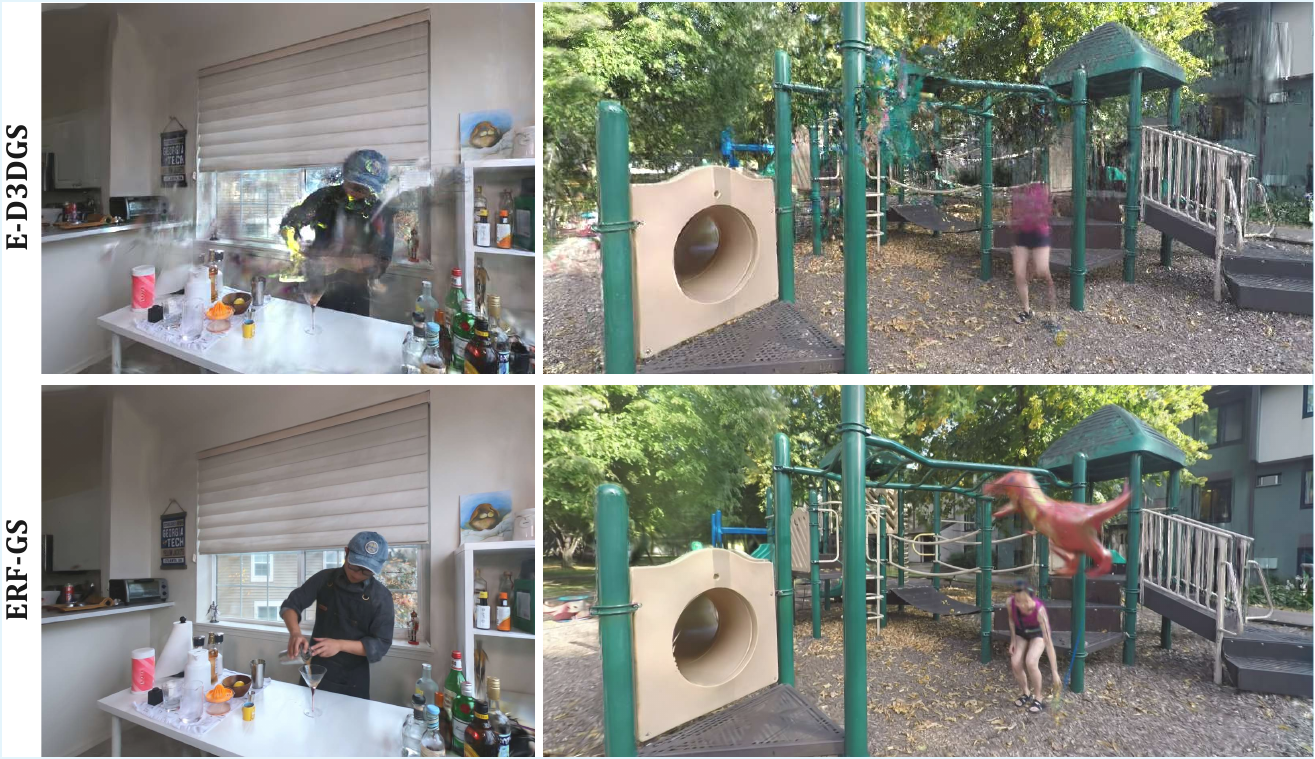}
   \caption{\textbf{Comparison of ERF-GS and E-D3DGS}: the latter exhibits artifacts such as color shift (left) and missing objects (right) due to inaccurate segmentation and unconstrained event-based supervision. 
   }
   \label{fig:ed3dgs}
\end{figure}

Despite the boom in event-based dynamic reconstruction methods, each of them employs a different protocol in terms of event simulation, viewpoint selection, event-RGB alignment and availability of off-the-shelf vision models, making it hard to compare between different baselines. Here we compare ERF-GS to E-D3DGS~\cite{xu2025event}, a recent and representative baseline, on Neu3D-dv and Nvidia-dv (see Table~\ref{tab:ed3dgs} and Fig.~\ref{fig:ed3dgs}).

Although E-D3DGS is designed for aligned event-RGB viewpoints, its core method is generalizable to our \textbf{-dv} settings. To make it runnable with 24GB GPU memory, we have to downsample the input images by 2$\times$2, essentially reducing the reconstruction difficulty. Nonetheless, we still observe it to exhibit severe artifacts. We attribute this to two key differences between our method and theirs: (a) E-D3DGS relies heavily on foreground-background segmentation estimated by foundation models, which can fail in complex real-world scenes and subsequently corrupt the whole reconstruction pipeline. (b) As for event-based supervision, E-D3DGS fails to disentangle it from Gaussian colors, which leads to color shifts and also disrupts the optimization of Gaussian shapes and motion. In contrast, these issues are promptly addressed by EARL and EDS in our ERF-GS framework.

\begin{table}[!t]
\caption{{Quantitative comparison of ERF-GS and E-D3DGS.} DPSNR$\uparrow$ and PSNR$\uparrow$ are reported.}
\centering
{
\begin{tabular}{l|cc}
\toprule
Dataset & \textbf{E-D3DGS} & \textbf{ERF-GS (Ours)}\\ 
\midrule
Neu3D-dv & 16.35 / 20.97 & \cellcolor{best}\textbf{26.65} / 26.29\\
Nvidia-dv & 17.10 / 20.37 & \cellcolor{best}\textbf{20.39} / 22.77\\
\bottomrule
\end{tabular}
}
\label{tab:ed3dgs}
\end{table}

\begin{figure*}[!t]
  \centering
     \includegraphics[width=\linewidth]{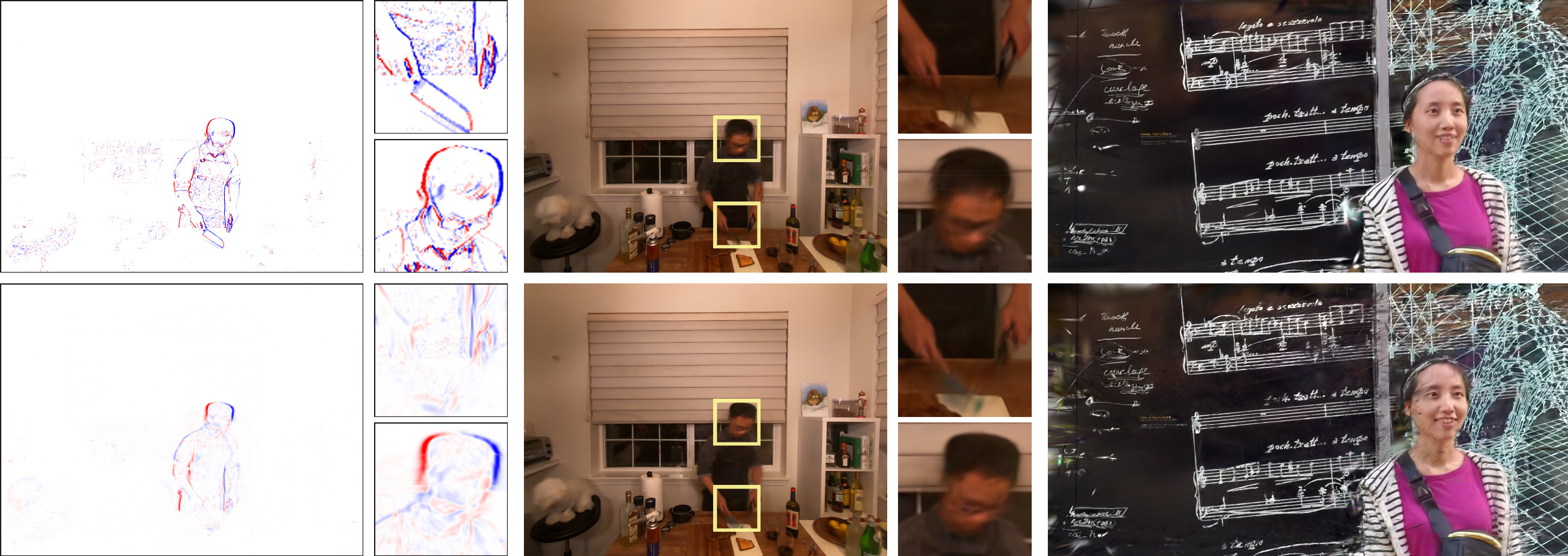}
   \caption{\emph{Failures using Neu3D-dv and Nvidia-dv.} Left: ground truth event frame ({above}) and estimated event frame using two rendered images ({below}) on `cut\_roasted\_beef'. Center: sample frames from baseline (above) and ERF-GS (below) on `cut\_roasted\_beef'. Right: sample frames from baseline (above) and ERF-GS (below) on `Dynamicface'.   
   In these cases, our method fails to learn scene dynamics around reflective surfaces and struggles with intensity changes in dark-colored regions. The resulting reconstruction quality in such regions is poor.
   }
   \label{fig:supp}
\end{figure*}

\section{Discussion and Conclusions}\label{sec:conclusion}
\subsection{Summary}
In this work, we have proposed ERF-GS, a novel event-RGB fusion framework for fast dynamic scene reconstruction. With EARL and EDS as its primary components, ERF-GS achieves event-based scene content learning without RGB supervision, and thus demonstrates significant performance improvements on challenge datasets with temporally subsampled and motion-blurred natural videos. 
Compared to existing event-based GS approaches, ERF-GS is particularly well-suited to real-world scenarios, where the alignment between modalities is difficult to attain and  RGB video quality is compromised by motion blur. As such, we expect ERF-GS to be the natural choice for reconstructing fast-moving natural 3D scenes with event-RGB inputs.

\subsection{Analysis of Failures}
Our solution for learning fast scene dynamics is not without flaws. Event activations, being proportional to changes in log-intensity, may exhibit larger errors in dark regions or when there are abrupt intensity changes. 
Here we analyze two cases of failure  that we encountered during experiments, using `cut\_roasted\_beef' from Neu3D-dv and `Dynamicface' from Nvidia-dv (see Fig.~\ref{fig:supp}). In the former scene, our ERF-GS falls behind the baseline by 1.08~dB in DPSNR and 0.19~dB in PSNR, while in the latter scene, the gaps are 0.18~dB and 0.35~dB, respectively. We attribute the performance drop to two main causes: reflective surfaces and dark regions.

Objects with reflective surfaces have long been a bugbear of GS-based methods~\cite{wu2024local}, and the challenge is even harder to resolve in dynamic scenes. In `cut\_roasted\_beef', both the knife and the tongs have reflective metal surfaces, and their visual appearances constantly change with motion, inducing color shifts (see Fig.~\ref{fig:supp}(above)) in event-based methods. In `Dynamicface', the background blackboard contains a large area of complex reflective patterns. While the board itself stays static throughout the video, changes in environmental light cause it to flicker and subsequently produce event activations that are not related to object motion. The ERF-GS model is misled by these events and creates floating Gaussians to model the nonexistent dynamics, which greatly impacts reconstruction quality (see Fig~\ref{fig:supp}(right)).

Since event activations are proportional to changes in the log-intensity of pixels, they are naturally more sensitive to low-light regions than RGB signals. While such a trait proves beneficial for downstream applications such as HDR imaging, it also makes the reconstruction of dark dynamic objects harder for ERF-GS, since a larger number of events means larger noise and sheerer loss landscapes. An example is provided by `cut\_roasted\_beef', when the actor lifts his head and causes an abrupt intensity change . As  Fig.~\ref{fig:supp}(below) shows, ERF-GS fails to learn the motion of his dark hair and therefore the shape of his head is distorted in the reconstructed video.

\subsection{Limitations and Future Work}
A core limitation of this work lies in the gap between simulated and real-world data. On the one hand, our simulated motion blur (see Fig.~\ref{fig:mbdv}) can appear unrealistic when object motion between two original RGB frames is too large. Possible mitigation strategies for this issue are to: (a) estimate optical flow for each frame with pre-trained models, construct a blur kernel for each pixel, and compute motion blur through convolution, or (b) interpolate between two frames with pre-trained models and take an average of the resulting denser set of images. However, both approaches would suffer from extended computation time and would be subject to inconsistencies between different frames due to the unavoidable imperfection of pre-trained models. Therefore, for now, we have resorted to a faster and more stable approach, which may sacrifice realism to some extent.

On the other hand, the domain gap between simulated and real-captured events is unavoidable. Although it is outside the scope of this paper to optimize  event simulation for realism, the efficacy of ERF-GS could indeed be better validated if more realistic simulators were available. Moreover, the temporal resolution of simulated event streams is limited by the frame rate of the original RGB videos. While we have attempted to replicate events' high-speed nature by subsampling RGB frames, we are still unable to simulate continuous event streams and impose event-based supervision at arbitrary times. Finally, the relatively large motion between consecutive video frames can disrupt the sparsity of event inputs, decreasing EARL's fidelity and inducing false intersections during EDS. 

Nevertheless, besides the simulation algorithm itself, we also believe that constructing a realistic training setup and relaxing the dependency of the reconstruction pipeline on assumptions such as viewpoint alignment can also help narrow this gap and lead to more practical event-RGB fusion frameworks. Our ERF-GS framework, designed and validated following this principle, reflects one of the first few works to tackle the lack of real-world event-RGB multiview data, which are too costly to collect at this stage of research.

Aware of the limitations of simulated data, our forthcoming research will focus on two main objectives: acquiring realistic multiview event-RGB datasets with fast scene dynamics, and enhancing our framework to fully exploit the distinct characteristics of real-world events.
To adapt our framework to the fields of robotics and AR/VR, another important stream of future research will involve generalizing it to a more constrained setting with few viewpoints, narrow fields-of-view, dynamic camera poses, and unknown camera trajectories. In our current setting, deploying ERF-GS on real-world hardware would pose engineering challenges such as synchronizing  over 10 different cameras and keeping them static while capturing dynamic scenes. We plan to incorporate existing advances in monocular 3D reconstruction and simultaneous optimization of 3D scene representation and camera poses to achieve this goal.

Moving forward, the generality of ERF-GS opens up new possibilities for its integration with pioneering works in the booming field of video reconstruction, which feature not only higher visual quality, but also advancements such as controllable and editable models, efficient learning pipelines, real-time streaming capability, smoother and invertible deformation representations, etc. We anticipate the synergy between events and Gaussian splatting to be further strengthened through exploring those possibilities.

{\small
\bibliographystyle{ieeenat_fullname}
\bibliography{references}

@String(CVPR= {IEEE Conf. Comput. Vis. Pattern Recog.})

@String(ICCV= {Int. Conf. Comput. Vis.})

@String(ICIP = {IEEE Int. Conf. Image Process.})

@inproceedings{hu2021v2e,
  title={v2e: From video frames to realistic DVS events},
  author={Hu, Yuhuang and Liu, Shih-Chii and Delbruck, Tobi},
  booktitle={Proceedings of the IEEE/CVF conference on computer vision and pattern recognition},
  pages={1312--1321},
  year={2021}
}

@article{park2021hypernerf,
  title={Hypernerf: A higher-dimensional representation for topologically varying neural radiance fields},
  author={Park, Keunhong and Sinha, Utkarsh and Hedman, Peter and Barron, Jonathan T and Bouaziz, Sofien and Goldman, Dan B and Martin-Brualla, Ricardo and Seitz, Steven M},
  journal={arXiv preprint arXiv:2106.13228},
  year={2021}
}

@article{mildenhall2021nerf,
  title={{NeRF}: Representing scenes as neural radiance fields for view synthesis},
  author={Mildenhall, Ben and Srinivasan, Pratul P and Tancik, Matthew and Barron, Jonathan T and Ramamoorthi, Ravi and Ng, Ren},
  journal={Communications of the ACM},
  volume={65},
  number={1},
  pages={99--106},
  year={2021},
  publisher={ACM New York, NY, USA}
}

@article{kerbl20233d,
  title={{3D} Gaussian Splatting for Real-Time Radiance Field Rendering.},
  author={Kerbl, Bernhard and Kopanas, Georgios and Leimk{\"u}hler, Thomas and Drettakis, George},
  journal={ACM Trans. Graph.},
  volume={42},
  number={4},
  pages={139--1},
  year={2023}
}

@article{wang2024nerf,
  title={{NeRF} in Robotics: A Survey},
  author={Wang, Guangming and Pan, Lei and Peng, Songyou and Liu, Shaohui and Xu, Chenfeng and Miao, Yanzi and Zhan, Wei and Tomizuka, Masayoshi and Pollefeys, Marc and Wang, Hesheng},
  journal={arXiv preprint arXiv:2405.01333},
  year={2024}
}

@article{ingale2021real,
  title={Real-time 3D reconstruction techniques applied in dynamic scenes: A systematic literature review},
  author={Ingale, Anupama K and others},
  journal={Computer Science Review},
  volume={39},
  pages={100338},
  year={2021},
  publisher={Elsevier}
}

@inproceedings{pumarola2021d,
  title={D-nerf: Neural radiance fields for dynamic scenes},
  author={Pumarola, Albert and Corona, Enric and Pons-Moll, Gerard and Moreno-Noguer, Francesc},
  booktitle={Proceedings of the IEEE/CVF Conference on Computer Vision and Pattern Recognition},
  pages={10318--10327},
  year={2021}
}

@inproceedings{li2023dynibar,
  title={Dynibar: Neural dynamic image-based rendering},
  author={Li, Zhengqi and Wang, Qianqian and Cole, Forrester and Tucker, Richard and Snavely, Noah},
  booktitle={Proceedings of the IEEE/CVF Conference on Computer Vision and Pattern Recognition},
  pages={4273--4284},
  year={2023}
}

@inproceedings{wu20244d,
  title={4d gaussian splatting for real-time dynamic scene rendering},
  author={Wu, Guanjun and Yi, Taoran and Fang, Jiemin and Xie, Lingxi and Zhang, Xiaopeng and Wei, Wei and Liu, Wenyu and Tian, Qi and Wang, Xinggang},
  booktitle={Proceedings of the IEEE/CVF Conference on Computer Vision and Pattern Recognition},
  pages={20310--20320},
  year={2024}
}

@inproceedings{li2024spacetime,
  title={Spacetime gaussian feature splatting for real-time dynamic view synthesis},
  author={Li, Zhan and Chen, Zhang and Li, Zhong and Xu, Yi},
  booktitle={Proceedings of the IEEE/CVF Conference on Computer Vision and Pattern Recognition},
  pages={8508--8520},
  year={2024}
}

@article{gallego2020event,
  title={Event-based vision: A survey},
  author={Gallego, Guillermo and Delbr{\"u}ck, Tobi and Orchard, Garrick and Bartolozzi, Chiara and Taba, Brian and Censi, Andrea and Leutenegger, Stefan and Davison, Andrew J and Conradt, J{\"o}rg and Daniilidis, Kostas and others},
  journal={IEEE transactions on pattern analysis and machine intelligence},
  volume={44},
  number={1},
  pages={154--180},
  year={2020},
  publisher={IEEE}
}

@inproceedings{deng2022voxel,
  title={A voxel graph cnn for object classification with event cameras},
  author={Deng, Yongjian and Chen, Hao and Liu, Hai and Li, Youfu},
  booktitle={Proceedings of the IEEE/CVF Conference on Computer Vision and Pattern Recognition},
  pages={1172--1181},
  year={2022}
}

@inproceedings{zhang2021object,
  title={Object tracking by jointly exploiting frame and event domain},
  author={Zhang, Jiqing and Yang, Xin and Fu, Yingkai and Wei, Xiaopeng and Yin, Baocai and Dong, Bo},
  booktitle={Proceedings of the IEEE/CVF International Conference on Computer Vision},
  pages={13043--13052},
  year={2021}
}

@inproceedings{messikommer2022multi,
  title={Multi-bracket high dynamic range imaging with event cameras},
  author={Messikommer, Nico and Georgoulis, Stamatios and Gehrig, Daniel and Tulyakov, Stepan and Erbach, Julius and Bochicchio, Alfredo and Li, Yuanyou and Scaramuzza, Davide},
  booktitle={Proceedings of the IEEE/CVF conference on computer vision and pattern recognition},
  pages={547--557},
  year={2022}
}

@inproceedings{monforte2020exploiting,
  title={Exploiting event cameras for spatio-temporal prediction of fast-changing trajectories},
  author={Monforte, Marco and Arriandiaga, Ander and Glover, Arren and Bartolozzi, Chiara},
  booktitle={2nd IEEE International Conference on Artificial Intelligence Circuits and Systems (AICAS)},
  pages={108--112},
  year={2020},
}

@inproceedings{wu2024ev,
  title={Ev-GS: Event-based gaussian splatting for efficient and accurate radiance field rendering},
  author={Wu, Jingqian and Zhu, Shuo and Wang, Chutian and Lam, Edmund Y},
  booktitle={IEEE 34th International Workshop on Machine Learning for Signal Processing (MLSP)},
  pages={1--6},
  year={2024},
}

@inproceedings{deguchi2024e2gs,
  title={{E2GS}: Event Enhanced Gaussian Splatting},
  author={Deguchi, Hiroyuki and Masuda, Mana and Nakabayashi, Takuya and Saito, Hideo},
  booktitle={IEEE International Conference on Image Processing (ICIP)},
  pages={1676--1682},
  year={2024}
}

@article{xiong2024event3dgs,
  title={{Event3DGS}: Event-based {3D} Gaussian Splatting for Fast Egomotion},
  author={Xiong, Tianyi and Wu, Jiayi and He, Botao and Fermuller, Cornelia and Aloimonos, Yiannis and Huang, Heng and Metzler, Christopher A},
  journal={arXiv preprint arXiv:2406.02972},
  year={2024}
}

@inproceedings{li2022neural,
  title={Neural 3d video synthesis from multi-view video},
  author={Li, Tianye and Slavcheva, Mira and Zollhoefer, Michael and Green, Simon and Lassner, Christoph and Kim, Changil and Schmidt, Tanner and Lovegrove, Steven and Goesele, Michael and Newcombe, Richard and others},
  booktitle={Proceedings of the IEEE/CVF Conference on Computer Vision and Pattern Recognition},
  pages={5521--5531},
  year={2022}
}

@inproceedings{scheerlinck2020fast,
  title={Fast image reconstruction with an event camera},
  author={Scheerlinck, Cedric and Rebecq, Henri and Gehrig, Daniel and Barnes, Nick and Mahony, Robert and Scaramuzza, Davide},
  booktitle={Proceedings of the IEEE/CVF Winter Conference on Applications of Computer Vision},
  pages={156--163},
  year={2020}
}

@inproceedings{gao2021dynamic,
  title={Dynamic view synthesis from dynamic monocular video},
  author={Gao, Chen and Saraf, Ayush and Kopf, Johannes and Huang, Jia-Bin},
  booktitle={Proceedings of the IEEE/CVF International Conference on Computer Vision},
  pages={5712--5721},
  year={2021}
}

@inproceedings{park2021nerfies,
  title={Nerfies: Deformable neural radiance fields},
  author={Park, Keunhong and Sinha, Utkarsh and Barron, Jonathan T and Bouaziz, Sofien and Goldman, Dan B and Seitz, Steven M and Martin-Brualla, Ricardo},
  booktitle={Proceedings of the IEEE/CVF International Conference on Computer Vision},
  pages={5865--5874},
  year={2021}
}

@inproceedings{newcombe2015dynamicfusion,
  title={Dynamicfusion: Reconstruction and tracking of non-rigid scenes in real-time},
  author={Newcombe, Richard A and Fox, Dieter and Seitz, Steven M},
  booktitle={Proceedings of the IEEE conference on computer vision and pattern recognition},
  pages={343--352},
  year={2015}
}

@inproceedings{bozic2020deepdeform,
  title={Deepdeform: Learning non-rigid rgb-d reconstruction with semi-supervised data},
  author={Bozic, Aljaz and Zollhofer, Michael and Theobalt, Christian and Nie{\ss}ner, Matthias},
  booktitle={Proceedings of the IEEE/CVF Conference on Computer Vision and Pattern Recognition},
  pages={7002--7012},
  year={2020}
}

@inproceedings{du2021neural,
  title={Neural radiance flow for 4d view synthesis and video processing},
  author={Du, Yilun and Zhang, Yinan and Yu, Hong-Xing and Tenenbaum, Joshua B and Wu, Jiajun},
  booktitle={IEEE/CVF International Conference on Computer Vision (ICCV)},
  pages={14304--14314},
  year={2021},
}

@inproceedings{xian2021space,
  title={Space-time neural irradiance fields for free-viewpoint video},
  author={Xian, Wenqi and Huang, Jia-Bin and Kopf, Johannes and Kim, Changil},
  booktitle={Proceedings of the IEEE/CVF conference on computer vision and pattern recognition},
  pages={9421--9431},
  year={2021}
}

@inproceedings{yu2024mip,
  title={Mip-splatting: Alias-free 3d gaussian splatting},
  author={Yu, Zehao and Chen, Anpei and Huang, Binbin and Sattler, Torsten and Geiger, Andreas},
  booktitle={Proceedings of the IEEE/CVF Conference on Computer Vision and Pattern Recognition},
  pages={19447--19456},
  year={2024}
}

@article{ren2024octree,
  title={Octree-gs: Towards consistent real-time rendering with lod-structured 3d gaussians},
  author={Ren, Kerui and Jiang, Lihan and Lu, Tao and Yu, Mulin and Xu, Linning and Ni, Zhangkai and Dai, Bo},
  journal={arXiv preprint arXiv:2403.17898},
  year={2024}
}

@article{luiten2023dynamic,
  title={Dynamic 3d gaussians: Tracking by persistent dynamic view synthesis},
  author={Luiten, Jonathon and Kopanas, Georgios and Leibe, Bastian and Ramanan, Deva},
  journal={arXiv preprint arXiv:2308.09713},
  year={2023}
}

@inproceedings{yang2024deformable,
  title={Deformable 3d gaussians for high-fidelity monocular dynamic scene reconstruction},
  author={Yang, Ziyi and Gao, Xinyu and Zhou, Wen and Jiao, Shaohui and Zhang, Yuqing and Jin, Xiaogang},
  booktitle={Proceedings of the IEEE/CVF Conference on Computer Vision and Pattern Recognition},
  pages={20331--20341},
  year={2024}
}

@inproceedings{cao2023hexplane,
  title={Hexplane: A fast representation for dynamic scenes},
  author={Cao, Ang and Johnson, Justin},
  booktitle={Proceedings of the IEEE/CVF Conference on Computer Vision and Pattern Recognition},
  pages={130--141},
  year={2023}
}

@article{gao2024gaussianflow,
  title={Gaussianflow: Splatting gaussian dynamics for 4d content creation},
  author={Gao, Quankai and Xu, Qiangeng and Cao, Zhe and Mildenhall, Ben and Ma, Wenchao and Chen, Le and Tang, Danhang and Neumann, Ulrich},
  journal={arXiv preprint arXiv:2403.12365},
  year={2024}
}

@inproceedings{huang2024sc,
  title={Sc-gs: Sparse-controlled gaussian splatting for editable dynamic scenes},
  author={Huang, Yi-Hua and Sun, Yang-Tian and Yang, Ziyi and Lyu, Xiaoyang and Cao, Yan-Pei and Qi, Xiaojuan},
  booktitle={Proceedings of the IEEE/CVF Conference on Computer Vision and Pattern Recognition},
  pages={4220--4230},
  year={2024}
}

@inproceedings{zhou2024drivinggaussian,
  title={Drivinggaussian: Composite gaussian splatting for surrounding dynamic autonomous driving scenes},
  author={Zhou, Xiaoyu and Lin, Zhiwei and Shan, Xiaojun and Wang, Yongtao and Sun, Deqing and Yang, Ming-Hsuan},
  booktitle={Proceedings of the IEEE/CVF Conference on Computer Vision and Pattern Recognition},
  pages={21634--21643},
  year={2024}
}

@article{wilson2021echo,
  title={Echo-reconstruction: Audio-augmented 3d scene reconstruction},
  author={Wilson, Justin and Rewkowski, Nicholas and Lin, Ming C and Fuchs, Henry},
  journal={arXiv preprint arXiv:2110.02405},
  year={2021}
}

@article{lichtsteiner2008128,
  title={A 128 times 128 120 dB 15 us latency asynchronous temporal contrast vision sensor},
  author={Lichtsteiner, Patrick and Posch, Christoph and Delbruck, Tobi},
  journal={IEEE journal of solid-state circuits},
  volume={43},
  number={2},
  pages={566--576},
  year={2008},
  publisher={IEEE}
}

@inproceedings{delbruck2010activity,
  title={Activity-driven, event-based vision sensors},
  author={Delbr{\"u}ck, Tobi and Linares-Barranco, Bernabe and Culurciello, Eugenio and Posch, Christoph},
  booktitle={Proceedings of IEEE international symposium on circuits and systems},
  pages={2426--2429},
  year={2010},
}

@article{rebecq2019high,
  title={High speed and high dynamic range video with an event camera},
  author={Rebecq, Henri and Ranftl, Ren{\'e} and Koltun, Vladlen and Scaramuzza, Davide},
  journal={IEEE transactions on pattern analysis and machine intelligence},
  volume={43},
  number={6},
  pages={1964--1980},
  year={2019},
  publisher={IEEE}
}

@article{ercan2024hypere2vid,
  title={Hypere2vid: Improving event-based video reconstruction via hypernetworks},
  author={Ercan, Burak and Eker, Onur and Saglam, Canberk and Erdem, Aykut and Erdem, Erkut},
  journal={IEEE Transactions on Image Processing},
  year={2024},
  publisher={IEEE}
}

@article{angelopoulos2021event,
  title={Event-Based Near-Eye Gaze Tracking Beyond 10,000 Hz},
  author={Angelopoulos, Anastasios N and Martel, Julien NP and Kohli, Amit P and Conradt, J{\"o}rg and Wetzstein, Gordon},
  journal={IEEE Transactions on Visualization and Computer Graphics},
  volume={27},
  number={5},
  pages={2577--2586},
  year={2021},
  publisher={IEEE}
}

@inproceedings{messikommer2023data,
  title={Data-driven feature tracking for event cameras},
  author={Messikommer, Nico and Fang, Carter and Gehrig, Mathias and Scaramuzza, Davide},
  booktitle={Proceedings of the IEEE/CVF Conference on Computer Vision and Pattern Recognition},
  pages={5642--5651},
  year={2023}
}

@inproceedings{alonso2019ev,
  title={EV-SegNet: Semantic segmentation for event-based cameras},
  author={Alonso, Inigo and Murillo, Ana C},
  booktitle={Proceedings of the IEEE/CVF Conference on Computer Vision and Pattern Recognition Workshops},
  pages={0--0},
  year={2019}
}

@inproceedings{gehrig2023recurrent,
  title={Recurrent vision transformers for object detection with event cameras},
  author={Gehrig, Mathias and Scaramuzza, Davide},
  booktitle={Proceedings of the IEEE/CVF conference on computer vision and pattern recognition},
  pages={13884--13893},
  year={2023}
}

@inproceedings{moreno2022visual,
  title={Visual event-based egocentric human action recognition},
  author={Moreno-Rodr{\'\i}guez, Francisco J and Traver, V Javier and Barranco, Francisco and Dimiccoli, Mariella and Pla, Filiberto},
  booktitle={Iberian Conference on Pattern Recognition and Image Analysis},
  pages={402--414},
  year={2022}
}

@inproceedings{gehrig2021raft,
  title={E-raft: Dense optical flow from event cameras},
  author={Gehrig, Mathias and Millh{\"a}usler, Mario and Gehrig, Daniel and Scaramuzza, Davide},
  booktitle={IEEE International Conference on 3D Vision (3DV)},
  pages={197--206},
  year={2021}
}

@article{wan2022learning,
  title={Learning dense and continuous optical flow from an event camera},
  author={Wan, Zhexiong and Dai, Yuchao and Mao, Yuxin},
  journal={IEEE Transactions on Image Processing},
  volume={31},
  pages={7237--7251},
  year={2022},
  publisher={IEEE}
}

@article{ieng2017event,
  title={Event-based 3D motion flow estimation using 4D spatio temporal subspaces properties},
  author={Ieng, Sio-Hoi and Carneiro, Jo{\~a}o and Benosman, Ryad B},
  journal={Frontiers in Neuroscience},
  volume={10},
  pages={596},
  year={2017},
  publisher={Frontiers Media SA}
}

@inproceedings{ma2023deformable,
  title={Deformable neural radiance fields using rgb and event cameras},
  author={Ma, Qi and Paudel, Danda Pani and Chhatkuli, Ajad and Van Gool, Luc},
  booktitle={Proceedings of the IEEE/CVF International Conference on Computer Vision},
  pages={3590--3600},
  year={2023}
}

@inproceedings{xu2025event,
  title={Event-boosted deformable 3d gaussians for dynamic scene reconstruction},
  author={Xu, Wenhao and Weng, Wenming and Zhang, Yueyi and Xu, Ruikang and Xiong, Zhiwei},
  booktitle={Proceedings of the IEEE/CVF International Conference on Computer Vision},
  pages={28334--28343},
  year={2025}
}

@inproceedings{liao2024ef,
    title={{EF}-3{DGS}: Event-Aided Free-Trajectory 3D Gaussian Splatting},
    author={Bohao Liao and Wei Zhai and Zengyu Wan and Zhixin Cheng and Wenfei Yang and Yang Cao and Tianzhu Zhang and Zheng-Jun Zha},
    booktitle={The Thirty-ninth Annual Conference on Neural Information Processing Systems},
    year={2025},
    url={https://openreview.net/forum?id=shFhW4zqd6}
}

@article{wang2024evggs,
  title={EvGGS: A Collaborative Learning Framework for Event-based Generalizable Gaussian Splatting},
  author={Wang, Jiaxu and He, Junhao and Zhang, Ziyi and Sun, Mingyuan and Sun, Jingkai and Xu, Renjing},
  journal={arXiv preprint arXiv:2405.14959},
  year={2024}
}

@article{rudnev2024dynamic,
  title={Dynamic EventNeRF: Reconstructing General Dynamic Scenes from Multi-view Event Cameras},
  author={Rudnev, Viktor and Fox, Gereon and Elgharib, Mohamed and Theobalt, Christian and Golyanik, Vladislav},
  journal={arXiv preprint arXiv:2412.06770},
  year={2024}
}

@inproceedings{yoon2020novel,
  title={Novel view synthesis of dynamic scenes with globally coherent depths from a monocular camera},
  author={Yoon, Jae Shin and Kim, Kihwan and Gallo, Orazio and Park, Hyun Soo and Kautz, Jan},
  booktitle={Proceedings of the IEEE/CVF Conference on Computer Vision and Pattern Recognition},
  pages={5336--5345},
  year={2020}
}

@article{ravi2024sam2,
  title={{SAM} 2: Segment Anything in Images and Videos},
  author={Ravi, Nikhila and Gabeur, Valentin and Hu, Yuan-Ting and Hu, Ronghang and Ryali, Chaitanya and Ma, Tengyu and Khedr, Haitham and R{\"a}dle, Roman and Rolland, Chloe and Gustafson, Laura and Mintun, Eric and Pan, Junting and Alwala, Kalyan Vasudev and Carion, Nicolas and Wu, Chao-Yuan and Girshick, Ross and Doll{\'a}r, Piotr and Feichtenhofer, Christoph},
  journal={arXiv preprint arXiv:2408.00714},
  url={https://arxiv.org/abs/2408.00714},
  year={2024}
}

@article{mueggler2017event,
  title={The event-camera dataset and simulator: Event-based data for pose estimation, visual odometry, and SLAM},
  author={Mueggler, Elias and Rebecq, Henri and Gallego, Guillermo and Delbruck, Tobi and Scaramuzza, Davide},
  journal={The International Journal of Robotics Research},
  volume={36},
  number={2},
  pages={142--149},
  year={2017},
  publisher={SAGE Publications Sage UK: London, England}
}

@inproceedings{wu2024local,
  title={Local Gaussian Density Mixtures for Unstructured Lumigraph Rendering},
  author={Wu, Xiuchao and Xu, Jiamin and Wang, Chi and Peng, Yifan and Huang, Qixing and Tompkin, James and Xu, Weiwei},
  booktitle={SIGGRAPH Asia 2024 Conference Papers},
  pages={1--11},
  year={2024}
}

@article{han2024event,
  title={Event-3dgs: Event-based 3d reconstruction using 3d gaussian splatting},
  author={Han, Haiqian and Li, Jianing and Wei, Henglu and Ji, Xiangyang},
  journal={Advances in Neural Information Processing Systems},
  volume={37},
  year={2024}
}

@inproceedings{schoenberger2016sfm,
    author={Sch\"{o}nberger, Johannes Lutz and Frahm, Jan-Michael},
    title={Structure-from-Motion Revisited},
    booktitle={Conference on Computer Vision and Pattern Recognition (CVPR)},
    pages={4104--4113},
    year={2016},
}

@inproceedings{louAllinFocusImagingEvent2023,
  title = {All-in-{{Focus Imaging}} from {{Event Focal Stack}}},
  booktitle = {{{IEEE}}/{{CVF Conference}} on {{Computer Vision}} and {{Pattern Recognition}} ({{CVPR}})},
  author = {Lou, Hanyue and Teng, Minggui and Yang, Yixin and Shi, Boxin},
  pages = {17366--17375},
  year={2023},
}

@article{lin2024embodied,
  title={Embodied neuromorphic synergy for lighting-robust machine vision to see in extreme bright},
  author={Lin, Shijie and Zheng, Guangze and Wang, Ziwei and Han, Ruihua and Xing, Wanli and Zhang, Zeqing and Peng, Yifan and Pan, Jia},
  journal={Nature Communications},
  volume={15},
  number={1},
  pages={10781},
  year={2024},
  publisher={Nature Publishing Group UK London}
}

@inproceedings{huang2024ev3dgs,
  title={{Ev3DGS}: Event Enhanced {3D} Gaussian Splatting from Blurry Images},
  author={Huang, Junwu and Wan, Zhexiong and Lu, Zhicheng and Zhu, Juanjuan and He, Mingyi and Dai, Yuchao},
  booktitle={IEEE Asia Pacific Signal and Information Processing Association Annual Summit and Conference (APSIPA ASC)},
  pages={1--6},
  year={2024}
}

@article{wan2025instance,
  title={Instance-Level Moving Object Segmentation from a Single Image with Events},
  author={Wan, Zhexiong and Fan, Bin and Hui, Le and Dai, Yuchao and Lee, Gim Hee},
  journal={International Journal of Computer Vision},
  pages={1--22},
  year={2025},
  publisher={Springer}
}

@article{duan2025eventaid,
  title={EventAid: Benchmarking event-aided image/video enhancement algorithms with real-captured hybrid dataset},
  author={Duan, Peiqi and Li, Boyu and Yang, Yixin and Lou, Hanyue and Teng, Minggui and Zhou, Xinyu and Ma, Yi and Shi, Boxin},
  journal={IEEE Transactions on Pattern Analysis and Machine Intelligence},
  year={2025},
  publisher={IEEE}
}

@article{wu2024sweepevgs,
  title={SweepEvGS: Event-Based 3D Gaussian Splatting for Macro and Micro Radiance Field Rendering from a Single Sweep},
  author={Wu, Jingqian and Zhu, Shuo and Wang, Chutian and Shi, Boxin and Lam, Edmund Y},
  journal={arXiv preprint arXiv:2412.11579},
  year={2024}
}

@article{zhu2018multivehicle,
  title={The multivehicle stereo event camera dataset: An event camera dataset for 3D perception},
  author={Zhu, Alex Zihao and Thakur, Dinesh and {\"O}zaslan, Tolga and Pfrommer, Bernd and Kumar, Vijay and Daniilidis, Kostas},
  journal={IEEE Robotics and Automation Letters},
  volume={3},
  number={3},
  pages={2032--2039},
  year={2018},
  publisher={IEEE}
}

@inproceedings{rebecq2018esim,
  title={Esim: an open event camera simulator},
  author={Rebecq, Henri and Gehrig, Daniel and Scaramuzza, Davide},
  booktitle={Conference on Robot Learning},
  pages={969--982},
  year={2018},
}

@inproceedings{yang2024dmit,
  title={DMiT: Deformable Mipmapped Tri-Plane Representation for Dynamic Scenes},
  author={Yang, Jing-Wen and Sun, Jia-Mu and Yang, Yong-Liang and Yang, Jie and Shan, Ying and Cao, Yan-Pei and Gao, Lin},
  booktitle={European Conference on Computer Vision},
  pages={436--453},
  year={2024},
}

@article{wu2024recent,
  title={Recent advances in 3d gaussian splatting},
  author={Wu, Tong and Yuan, Yu-Jie and Zhang, Ling-Xiao and Yang, Jie and Cao, Yan-Pei and Yan, Ling-Qi and Gao, Lin},
  journal={Computational Visual Media},
  volume={10},
  number={4},
  pages={613--642},
  year={2024},
  publisher={TUP}
}

@article{peng2025gaussian,
  title={Gaussian-plus-SDF SLAM: High-fidelity 3D reconstruction at 150+ fps},
  author={Peng, Zhexi and Zhou, Kun and Shao, Tianjia},
  journal={Computational Visual Media},
  year={2025},
  publisher={TUP}
}

@article{hou2024causal,
  title={A causal convolutional neural network for multi-subject motion modeling and generation},
  author={Hou, Shuaiying and Wang, Congyi and Zhuang, Wenlin and Chen, Yu and Wang, Yangang and Bao, Hujun and Chai, Jinxiang and Xu, Weiwei},
  journal={Computational Visual Media},
  volume={10},
  number={1},
  pages={45--59},
  year={2024},
  publisher={Springer}
}
}

\end{document}